\documentclass[conference]{IEEEtran}
\IEEEoverridecommandlockouts
\usepackage{graphicx} % Required for inserting images
\usepackage{cite}
\usepackage{amsmath,amssymb,amsfonts}
\usepackage{algorithmic}
\usepackage{graphicx}
\usepackage{textcomp}
\usepackage{xcolor}

\newcommand{\eat}[1]{}
\usepackage[square,numbers]{natbib} % has a nice set of citation styles and commands
\usepackage{mathtools} % amsmath with fixes and additions
\usepackage{booktabs} % commands to create good-looking tables
\usepackage{tikz} % nice language for creating drawings and diagrams
\usepackage{array}
\usepackage{amsmath}
\usepackage{amsthm}
\usepackage{amssymb}
\usepackage{subfigure}
\usepackage{xspace}
\usepackage{paralist}
\usepackage{multirow}
\usepackage[ruled,vlined,oldcommands]{algorithm2e}
\usepackage{url}
\usepackage{mathabx}
\usepackage{fancyhdr}
\usepackage{enumitem}

 \newtheorem{definition}{Definition}

\def\BibTeX{{\rm B\kern-.05em{\sc i\kern-.025em b}\kern-.08em
    T\kern-.1667em\lower.7ex\hbox{E}\kern-.125emX}}

\begin{document}

\title{Disentangling Fine-Tuning from Pre-Training in Visual Captioning with Hybrid Markov Logic\thanks{This research was supported by NSF award \#2008812. The opinions, findings, and results are solely the authors' and do not reflect those of the funding agencies.}}

\author{\IEEEauthorblockN{1\textsuperscript{st} Monika Shah}
% \IEEEauthorblockA{\textit{Computer Science} \\
\textit{University of Memphis}\\
% Memphis, USA \\
mshah2@memphis.edu
\and
\IEEEauthorblockN{2\textsuperscript{nd} Somdeb Sarkhel}
\textit{Adobe Research}\\
% San Jose, USA \\
sarkhel@adobe.com
\and
\IEEEauthorblockN{3\textsuperscript{rd} Deepak Venugopal}
% \IEEEauthorblockA{\textit{Computer Science} \\
\textit{University of Memphis}\\
% Memphis, USA \\
dvngopal@memphis.edu}

\maketitle

\begin{abstract}

Multimodal systems have highly complex processing pipelines and are pretrained over large datasets before being fine-tuned for specific tasks such as visual captioning. However, it becomes hard to disentangle what the model learns during the fine-tuning process from what it already knows due to its pretraining. In this work, we learn a probabilistic model using Hybrid Markov Logic Networks (HMLNs) over the training examples by relating symbolic knowledge (extracted from the caption) with visual features (extracted from the image). For a generated caption, we quantify the influence of training examples based on the HMLN distribution using probabilistic inference. We evaluate two types of inference procedures on the MSCOCO dataset for different types of captioning models. Our results show that for BLIP2 (a model that uses a LLM), the fine-tuning may have smaller influence on the knowledge the model has acquired since it may have more general knowledge to perform visual captioning as compared to models that do not use a LLM.

\end{abstract}


\section{Introduction}

%The performance of deep neural networks (DNNs) on multimodal tasks such as visual captioning has improved tremendously over the past decade. However, evaluating the 

Image captioning is arguably one of the most popular multimodal AI applications that integrates language processing, visual understanding and knowledge representation. Well-known captioning models~\cite{li2023blip, pan2020x} generate captions that are often similar to {\em ground-truth} captions written by humans. The degree of similarity is typically quantified using scores such as BLEU, CiDer, ROGUE and METEOR~\cite{papineni2002bleu, denkowski2014meteor, anderson2016spice}. Other reference-free metrics~\cite{madhyastha2019vifidel, hessel2021clipscore} attempt to analyze if the generated text describes key features in the image. A popular approach is CLIPScore~\cite{hessel2021clipscore} which embeds the generated caption and the test image into a common embedding-space using the CLIP model and evaluates the caption based on the similarity between the text and image embeddings. While all of the aforementioned methods tell us if a captioning model generates high-quality captions, they are not designed to analyze the influence of the data distribution on the outputs observed from the model.

\begin{figure}
    \centering
    \includegraphics[scale=0.38]{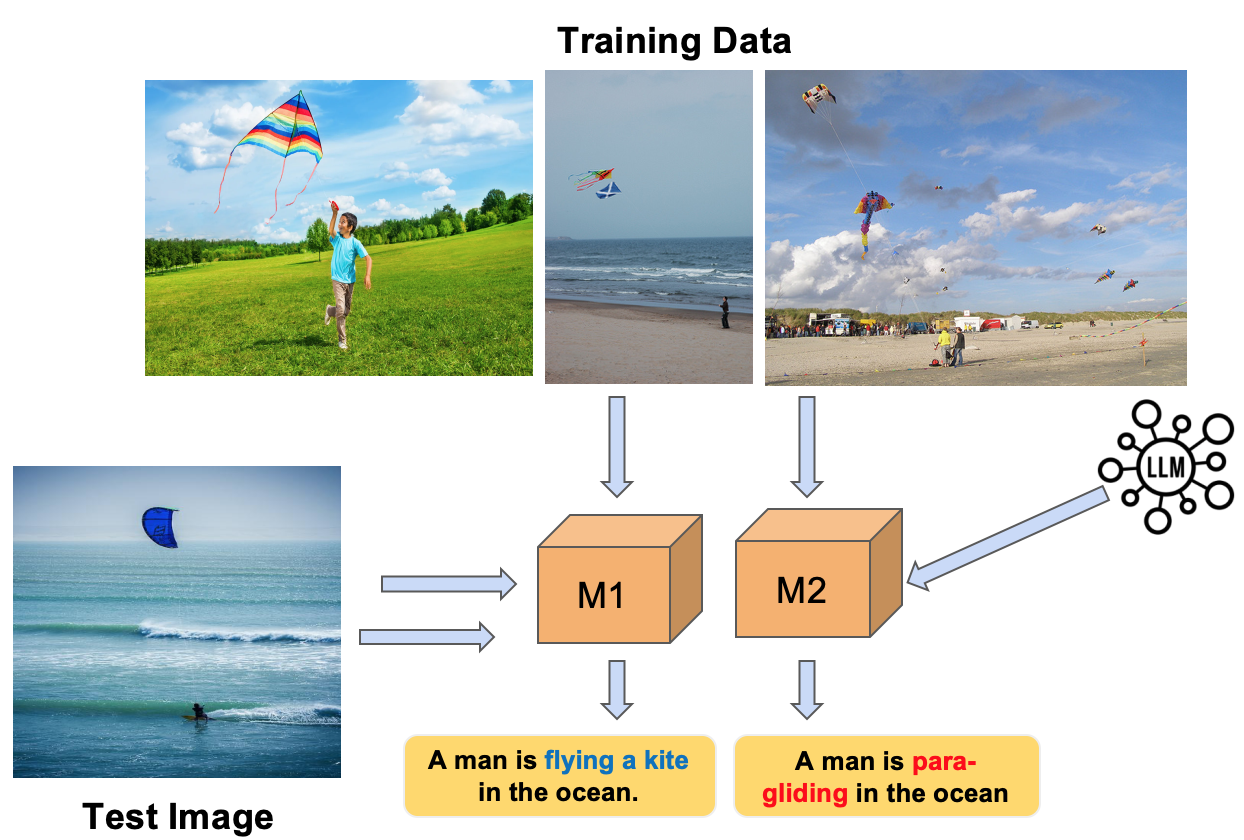}
    \caption{Illustrative example where M2 is pretrained with an LLM while M1 is not. For M1, the source of its knowledge in generating the caption can be traced in the training examples while for M2 it is harder since it may have more general knowledge from pretraining.}
    \label{fig:introex}
\end{figure}

%Particularly, almost all captioning models are often {\em pre-trained} over very large, general datasets before they are fine-tuned with examples from a more specific target distribution. Therefore, if we were to analyze concept learning in the machine learning models that learns by observation of samples from the target distribution, we need a framework to {\em disentangle} outputs observed from the model based on the training distribution. Without such an analysis, it is hard to verify how training examples contribute to the model's learning. For instance, suppose we consider the example in Fig.~\ref{fig:introex}, the model $M2$ uses pretraining sources and generates a novel concept in the caption ({\em para-gliding}) which makes it less explainable w.r.t the training distribution.

Almost all captioning models are often {\em pre-trained} over very large, general datasets before they are fine-tuned with examples from a more specific target distribution. In this work we want to {\em disentangle what the model learns from examples from a training distribution used for fine-tuning from more general features it has learned during pre-training}. Such disentanglement can help explain if fine-tuning adds/removes/modifies knowledge acquired by the pre-trained model. For instance, suppose we consider the example in Fig.~\ref{fig:introex}, the model $M1$ generates the caption with {\em kite-flying} which can be observed in the training examples, but the VLLM model generates a caption with {\em para-gliding} that cannot be traced to examples in the training dataset used for fine-tuning.

%For instance, suppose we consider the example in Fig.~\ref{}, the caption that learns from the training data generates a distinct caption as compared to the caption that utilizes sources other than the training data to generate a caption. 
%the concepts that need to be learned to generate the caption in  can be explained by samples from the distribution shown in the figure. 
%but still be considered as the {\em correct model} which is problematic. 
%pre-training of models over large datasets 
%To explain caption generation, we {\em disentangle} learning using training examples from {\em black-box} sources of learning. 
To do this, we interpret the generated caption as a sample drawn from the training distribution. We learn this distribution based on concepts seen by the model in the training examples. To do this, we use a language called Hybrid Markov Logic Networks (HMLNs)~\cite{wang2008hybrid}, where we can define a unified distribution over symbolic constraints combined with real-valued functions.
This allows us to define a distribution over text and visual modalities, i.e., we combine symbolic concepts (from text) with real-valued functions over visual features (from the image). To explain how a model learns concepts to generate a caption from this distribution, we formalize two inference tasks.
%Further, we develop two inference methods to explain the generation of captions. 
The first one is {\em abductive inference} which is formalized as computing the Max a Posteriori (MAP) estimate. Here, conditioned on the visual features in the image, we compute if the generated caption has the maximum probability. To do this, we add constraints between concepts in the generated caption and those in the true distribution using semantic equivalence and compute the approximate MAP solution using an off-the-shelf Mixed-Integer Linear Programming (MILP) solver.
%In our case, this explanation is an  to concepts in the generated caption. To performing MAP, we develop 
%In our second task, we sample examples from the training data based on the distributio. Specifically, we formalize bias in the generated caption using importance weights over training examples. We then estimate probabilities for examples related to the generated caption using these weights and select examples that have contrasting probabilities as an explanation of the model's learning. 
In our second inference task, we infer the probability of training examples conditioned on the generated caption. A larger probability indicates that the training instance has greater relevance to the generated caption. 

We evaluate both inference methods on the MSCOCO dataset where we evaluate varied types of captioning models such as SGAE~\cite{yang2019auto}, transformer-based models~\cite{huang2019attention, pan2020x, cornia2020meshed} and BLIP2 (a Visual Large Language Model (VLLM) based model)~\cite{li2023blip}. Our results for MAP inference shows that captions generated using BLIP2 have smaller MAP probabilities compared to the other methods. We also design an Amazon Mechanical Turk (AMT) study where we ask users if contrastive training examples that have high/low probabilities w.r.t a generated caption matched with their own perception of how the model acquired knowledge for caption generation. For the case of BLIP2 generated captions, users found that the contrastive examples were less interpretable as compared to other methods. The results seem to indicate that fine-tuning may have smaller influence in a model such as BLIP2 (since it has more general knowledge through LLMs) compared to other models that did not use LLMs.

\section{Related Work}

There are several methods that relate visual features with the caption text to analyze the performance of captioning systems. Specifically, CLIPScore~\cite{hessel2021clipscore} uses the well-known CLIP model to embed the image and its corresponding caption to measure coherence between the two.
Similarly, VIFIDEL~\cite{madhyastha2019vifidel} measures the visual fidelity by comparing a representation of the generated caption with the visual content. CLAIR~\cite{chan2023clair} prompts a Large Language Model to score generated captions.

In~\cite{shalev2022baseline}, a probabilistic model is used to detect out of distribution (OOD) images in image captioning. In particular, several manifestations of OOD instances such as those containing novel objects, corrupted or anomalous images, low-quality images, etc. are identified with this approach.  In~\cite{monika&al22}, an approach to evaluate captioning models was developed from distributions learned using Markov Logic Networks (MLNs). However, unlike HMLNs, MLNs do not allow us to specify real-valued functions over visual features. More recently, a more general approach for verification of real-valued functions over embeddings by combining them with symbolic knowledge using HMLNs was proposed in~\cite{shakya&al23}. However, unlike our proposed approach, the previous approaches were not designed to disentangle fine-tuning from pre-training. Our approach is also related more generally to the work in \cite{grover2019bias} which proposed an approach to quantify learning bias in a generative model. 

The area of knowledge disentanglement has been studied in different contexts. In ~\cite{an2024knowledge}, an approach is proposed for disentangled knowledge acquisition in visual question answering, where the question is decomposed into sub-questions and knowledge required for each sub-question is obtained from different sources. Training Data Attribution methods have been proposed as a way to trace factual knowledge back to the training examples. This allows us to determine where a Large Language Model (LLM) acquired its knowledge from and increases trust in the model~\cite{akyurek2022towards}. In ~\cite{wang2024generalizationvsmemorizationtracing}, an approach orthogonal to ours is proposed where the idea is to trace LLM capabilities back to pretraining data. This helps us understand the extent to which memorization from large-scale examples helps an LLM achieve impressive performance and the degree to which they can actually generalize.

\section{Background}
Here, we provide a brief background on HMLNs and inference using Gibbs sampling that we utilize in our approach.

\subsection{Hybrid Markov Logic Networks}
A HMLN is a set of pairs $(f_i,\theta_i)$, where $f_i$ is a First-Order Logic (FOL) formula~\cite{genesereth2013introduction} and $\theta_i$ is a weight associated with that formula to represent its uncertainty. $f_i$ is called a hybrid formula if it contains both continuous functions along with symbolic/discrete terms. Given a domain of constants that can be substituted for variables in the formulas, we can {\em ground} a formula by substituting all its variables with constants from their respective {\em domains}. A predicate symbol specifies a relation over terms. A {\em ground predicate} is one where all the variables in its arguments have been replaced by constants/objects from their respective domains. 

In the HMLN distribution, a ground predicate is considered as a random variable. Ground predicates that are discrete/symbolic take values 0/1 and numeric terms take continuous values. We use the terms ground predicate or ground atom interchangeably. The value of a ground formula is equal to 0/1 if it contains no numeric terms while a ground formula with numeric terms has a continuous value. 

We can represent the distribution of a HMLN as a product of potential functions, where each potential function specifies a ground formula. Specifically, the distribution takes a log-linear form where each grounding has a value equal to $e^{\theta\times s}$, where $\theta$ is the weight of the formula that the potential represents and $s$ is the value of the formula (for a purely symbolic formula this is 0/1). We can specify the product over all potentials in the distribution as follows.
\begin{equation}
    P({\bf X}={\bf x})=\frac{1}{Z}\exp\left(\sum_i\theta_is_i({\bf x})\right)
\end{equation}
where ${\bf x}$ is a {\em world}, i.e., an assignment to the variables defined in the HMLN distribution, $s_i({\bf x})$ is the value of the $i$-th formula given the world ${\bf x}$, $Z$ is the normalization constant also called as the partition function which is intractable to compute since it is a summation over all possible worlds.

\subsection{Gibbs Sampling}

We can estimate expected values from the HMLN distribution using Gibbs sampling~\cite{geman1984stochastic}, a widely used Markov Chain Monte Carlo (MCMC) algorithm. Specifically, we generate samples from the distribution $P({\bf y}|{\bf X})$ as follows.
%We use Gibbs sampling~\cite{geman1984stochastic} to generate samples from $P({\bf Y}|{\bf X})$. 
We start with a random assignment to all non-evidence ground predicates ${\bf y}$. In each iteration, we sample the assignment to a ground predicate $y\in{\bf y}$ from a conditional distribution $P(y|{\bf X},MB(y))$, where $MB(y)$ denotes the Markov Blanket of $y$. $MB(y)$ is the set of ground predicates connected to $y$ via a formula in the HMLN. Conditioning on $MB(y)$ renders $y$ conditionally independent of all the other variables in the HMLN and therefore, computing the conditional distribution at each iteration is very efficient. Typically, the sampler is allowed to run for a few iterations (called the {\em burnin} time) to allow it to {\em mix} which ensures that it forgets its initialization and starts to collect samples from the target distribution. We can estimate any expectation from samples drawn from the mixed sampler and as the number of samples go to $\infty$, the expectations converge to their true values.

\section{Approach}
\subsection{Notation}
To formalize our approach, we use the following notation. We use upper-case letters ($X$, $Y$, etc.) to denote variables which in HMLNs are ground predicates. We use lower-case ($x$,$y$, etc.) to denote assignments on the variables. We use bold-face fonts for upper-case letters to denote multiple variables (${\bf X}$, ${\bf Y}$, etc.) and bold-face fonts on lower-case letters to denotes assignments (or a sample) over these variables  (${\bf x}$, ${\bf y}$, etc.). 
%We use calligraphic fonts ($\mathcal{X}$, $\mathcal{I}$, etc.) to denote examples/instances in the training or test set.
\subsection{HMLN Structure}

%We represent templates of hybrid formulas as properties over relations within a caption.
%We define the HMLN structure using templates of hybrid formulas (with symbolic and continuous terms). Note that the HMLN framework is flexible enough to represent arbitrary First Order Logic (FOL) [] structures. 
We define the HMLN structure using templates of hybrid formulas. Note that the HMLN framework is flexible enough to represent arbitrary FOL structures. However, in our case, we limit the HMLN to explainable structures and therefore, similar to the approach in ~\cite{wang2008hybrid}, we pre-define templates and then slot-fill the templates from training data.
\begin{definition}
\label{def:1}
    A conjunctive feature ($\mathcal{C}$) is a hybrid formula of the form, $f_c(X_i,\ldots X_k)$ $\times$ $(X_1\wedge X_2\ldots X_k)$, where $X_i$ and $X_{i+1}$ are ground predicates that share at least one common term and $f_c()$ is a real-valued function, and $\times$ indicates that we multiply the real-term with the symbolic formula.
\end{definition}
\begin{definition}
\label{def:2}
    An XOR feature ($\mathcal{I}$) is defined as the XOR over a pair of ground predicates, i.e., $f_r(X_i,X_j)$ $\times$ $(X_i\wedge \neg X_j)$ $\vee$ $(X_j\wedge \neg X_i)$ where $X_i,X_j$ share at least one variable between them and $f_r()$ is a real-valued function.
\end{definition}
The $\mathcal{C}$ property connects ground predicates that occur together within captions, i.e., concepts that are related to each other. The $\mathcal{I}$ property encodes our reasoning that observing a predicate explains away the existence of the other predicate. Specifically, we can write the symbolic part of the property as $(X_i\Rightarrow \neg X_j)$ $\wedge$ $(X_j\Rightarrow \neg X_i)$ if we ignore the case where the head of the implications are false. 
%The (soft) equality over the functions is a continuous approximation of one predicate explaining away the other predicate. 

% we relate the symbolic representation with the visual representation in the training data corresponding to the potential function of $f$. 

The training data $\mathcal{D}$ consists of image and caption pairs (where the caption is a human-written). For each image in the training data, we select its caption and extract ground predicates for this captions using a textual scene graph parser~\cite{schuster2015generating}. 
We slot-fill the $\mathcal{C}$ template by chaining predicates with shared objects in the same image. We limit the size of the chain to a maximum of two predicates since beyond this, the chains seem to lack coherence. For each $\mathcal{C}$ formula, we also fill its corresponding $\mathcal{I}$ template. 

%Thus, a pair of predicates with shared objects extracted from $D$ $\in$ $\mathcal{D}$ are grounded as two pairwise potential functions (corresponding to the $\mathcal{C}$ and $\mathcal{I}$ formulas) in the HMLN distribution.

\noindent{\bf Real-Valued Terms.} The real-valued terms in definitions \ref{def:1} and \ref{def:2} connect the visual and text modalities. Specifically, let $X_1,X_2$ represent two ground predicates in a formula that was extracted from $D$ $\in$ $\mathcal{D}$. We use CLIP~\cite{radford2021learning} to embed the image in $D$ and the predicates $X_1,X_2$ into a common embedding-space. Let $g(X_1)$, $g(X_2)$ represent the cosine similarities between the image embedding and each of the predicate embeddings respectively. Following the semantics defined in ~\cite{wang2008hybrid}, the real-valued terms in the $\mathcal{I}$ and $\mathcal{C}$ potentials are expressed as follows.
\begin{definition}
    $$f_c(X_1,X_2)=\min\{-\log\sigma(\epsilon-g(X_1)),-\log\sigma(\epsilon-g(X_2))\}$$
where $\sigma$ represents the sigmoid function and $\epsilon$ is a constant. In terms of HMLN semantics, the soft inequality $g(X_1)\geq\epsilon$ is represented by the negative log-sigmoid of $(\epsilon-g(X_1))$. Thus, if the visual representation agrees with the symbolic predicates resulting in larger values of $g(X_1),g(X_2)$, the function assigns a higher value to the formula.    
\end{definition}
\begin{definition}
$$f_r(X_1,X_2) = -(g(X_1)- g(X_2))^2$$
We can show that if the weight of the formula is $\theta$,  $f_r(\cdot)$ is a Gaussian penalty with standard deviation $\sqrt{1/2\theta}$. Larger differences between $g(X_1)$ and $g(X_2)$ imposes a greater penalty on the $\mathcal{I}$ formulas and thus assigns a smaller value to the formula.    
\end{definition}

Note that Lukasiewicz approximations for continuous logic~\cite{bach2017hinge} extends Boolean logic to the continuous case where each variable is allowed to take a continuous value in the interval $[0,1]$. The Lukasiewicz t-norm and t-co-norm define $\wedge$ and $\vee$ operators for continuous logic. The $\mathcal{C}$ and $\mathcal{I}$ values can be considered as variants of this approximation. Specifically, for the $\mathcal{C}$ value, we consider the minimum value (instead of the sum) since we want to penalize a formula even if one of its predicates is a poor descriptor for the image. Similarly, for the $\mathcal{I}$ value, we use the Gaussian penalty instead of a simple difference between the values.

%where the potential values are determined by the learned parameters and the values of their $\mathcal{C}$, $\mathcal{I}$ properties. 

%An example for potential functions is illustrated in Fig.~\ref{fig:virt}.

\subsection{Learning}

 %Next, we learn the parameters for the potentials using max-likelihood estimation.

We parameterize the HMLN structure using max-likelihood estimation. However, note that in our case learning the weights for all the potential functions jointly is infeasible since the number of groundings can be extremely large.
%Note that in a statistical relational model, we assume that all the training instances are interconnected.
%Typically, to learn parameters of the HMLN (or other statistical relational models), we  maximize the likelihood over the full training data. 
%In other words, we ground the full HMLN and compute the weights for all potential functions. 
%However, in our case the ground HMLN is extremely large which makes learning infeasible. 
Further, to perform inference on a specific query (in our case the output produced by a model for a single test instance) note that only a small number of groundings will be relevant to the query since we can assume that the other groundings will have a value equal to 0 and thus do not influence the inference results~\cite{domingos2009markov}. Therefore, we instead develop an {\em inference guided} parameterization where we learn parameters during inference time on a subset of the HMLN formula groundings that are relevant to the inference query. Inference based learning has also been utilized to scale-up learning in large graphical models~\cite{chechetka2011query}. To do this, we identify and localize objects in the image using Faster R-CNN~\cite{ren2015faster} and learn query-specific parameters for potential functions that contain these objects. 

%However, in our case this is not desirable due to two reasons. 

%First, this will add irrelevant groundings making it infeasible to learn the parameters. 

%Second, a single parameterization may

%when we want to generate fine-grained explanations specific to a generated caption. Therefore, we instead develop an {\em inference guided} parameterization where we learn parameters that are relevant to an inference query. Some related approaches such as ~\cite{chechetka2011query}, have explored similar learning methods. 

%Here, we {\em normalize} the HMLN structure w.r.t 

%Specifically, only the potentials that are related to the inference query 

%are relevant to the test image and the other groundings by default have a value equal to 0 (i.e., we can ignore these groundings). To do this, we identify and localize objects in the image using Faster R-CNN~\cite{ren2015faster} and learn query-specific parameters for groundings that contain these objects. 

%the test instance that is being explained. In the normal form, we make a closed world assumption where given a test image and the generated caption, only the groundings that contain objects that are identified in the image are relevant to the test image and the other groundings by default have a value equal to 0 (i.e., we can ignore these groundings). To do this, we identify and localize objects in the image using Faster R-CNN~\cite{ren2015faster} and learn query-specific parameters for groundings that contain these objects. 

\noindent{\bf Parameterization.} 
The log-likelihood function for the HMLN is as follows.
\begin{equation}
\label{eq:gen}
    \ell(\Theta:\mathcal{D})=\log\frac{1}{Z(\Theta)}\exp\left\{\sum_{i=1}^k\theta_i\mathbb{E}_{\mathcal{D}}[\mathcal{C}_i(\mathcal{X})+\mathcal{I}_i(\mathcal{X})]\right\}
\end{equation}

where $\Theta$ is the set of parameters to be learned, $Z(\Theta)$ is the normalization constant, $\mathbb{E}_{\mathcal{D}}[\mathcal{C}_i(\mathcal{X})+\mathcal{I}_i(\mathcal{X})]$ is the empirical expectation of the values in the $i$-th conjunctive and XOR features. Specifically, this is computed as the average value of the features over the dataset $\mathcal{D}$.
Since the model is log-linear, the gradient of $\ell(\Theta:\mathcal{D})$ w.r.t to the $i$-th weight is as follows~\cite{koller&friedman09}.
$$\frac{\partial(\ell(\Theta:\mathcal{D}))}{\partial{\theta_i}}=\mathbb{E}_{\mathcal{D}}[\mathcal{C}_i(\mathcal{X})+\mathcal{I}_i(\mathcal{X})]-\mathbb{E}_{\Theta}[\mathcal{C}_i(\mathcal{X})+\mathcal{I}_i(\mathcal{X})]$$

Note that each feature consists of both real and symbolic terms. Eq.~\ref{eq:gen} corresponds to generative learning since we jointly optimize over the both terms. However, in our case, models use the image to learn relationships described in a caption. Therefore, we use discriminative learning where we condition on the real-valued terms. Let ${\bf x}[m]$ denote the real-valued terms and ${\bf y}[m]$ the symbolic terms for the $m$-th instance in the training data. The gradient of the conditional log-likelihood (CLL) is as follows.
\begin{align}
\label{eq:grad}
    \frac{\partial(\ell(\Theta:\mathcal{D}))}{\partial{\theta_i}}=\sum_{m=1}^M(\mathcal{C}_i({\bf y}[m],{\bf x}[m])+\mathcal{I}_i({\bf y}[m],{\bf x}[m]))-\nonumber\\
    \mathbb{E}_{\Theta}[\mathcal{C}_i({\bf y}[m]|{\bf x}[m])+\mathcal{I}_i({\bf y}[m]|{\bf x}[m])]
\end{align}
Though the CLL is concave and thus has a globally optimal parameterization, there is no analytical closed form solution. Therefore, we use iterative methods such as gradient ascent to maximize the CLL. However, this leads to a complication. Note that to compute the expected value of a feature, we need to compute the exact marginal probability of each feature which is intractable since the normalization constant is unknown. Therefore, we use contrastive divergence~\cite{hinton2002training} to approximate the expectation. Specifically, the idea is to draw samples from the distribution using the weights estimated thus far and estimate the expected value of a feature as a Monte-Carlo estimate over the samples. Note that since we need only the approximate gradient direction, a few samples are sufficient to perform the weight update efficiently. To do this, we use Gibbs sampling to generate the samples and compute the expected value of a feature as the average value computed over the samples. Note that we can update the weights of all the features in parallel by jointly sampling ${\bf y}[m]$ given ${\bf x}[m]$ for all $m$. Algorithm~\ref{alg:wlearning} summarizes our parameter learning approach.

%~\cite{geman1984stochastic},  a well-known Markov Chain Monte Carlo (MCMC) sampling approach. 

%We compute the 

%In particular, we use Gibbs sampling~\cite{geman1984stochastic} to  compute the expected value of a feature as the average value of the feature over all the samples.
%CONDITIONAL LOG-LIKELIHOOD

%using a standard gradient ascent method. The
%\mathcal{C}_i({\bf d}_i$ is the value of the $i$-th conjunctive feature and $\mathcal{I}_i({\bf d}_i)$ is the value of the $i$-th XOR feature.
%$\phi_i({\bf x})$ $=$ $\exp(w_i*(\mathcal{I}_i({\bf x})+\mathcal{C}_i({\bf x})))$

%where $\mathcal{I}_i({\bf x})$ is the value of the $\mathcal{I}_i$ property

%$f$ is the symbolic formula associated with weight $w_f$ and $\phi_j\sim f$ denotes that the potential $\phi_j$ shares the symbolic formula $f$, i.e., $\phi_j$ is a grounding of $f$. Note that $\forall$ $\phi_j\sim f$, $w_f$ is a shared weight. $\mathcal{I}_j({\bf x})$, $\mathcal{C}_j({\bf x})$ are the $\mathcal{I}$, $\mathcal{C}$ values in $\phi_j$.

\begin{algorithm}[!t]{
\small
\linesnumbered
\caption{Parameter Learning}\label{alg:wlearning}
\KwIn{Query specific HMLN structure $\mathcal{H}$, Training data $\mathcal{D}$, learning rate $\eta$}
\KwOut{Learned weights $\{\theta_i\}_{i=1}^k$}
\tcp{Contrastive Divergence}
Initialize weights $\{\theta_i^{(0)}\}$\\ 
\For{$t$ from 1 to $T$}
{
$\{{\bf x}^{(t)},{\bf y}^{(t)}\}$ $=$ Sample from $\mathcal{H}$ parameterized with $\{\theta_i^{(t-1)}\}_{i=1}^k$ using Gibbs sampling\\
\For{each weight $\theta_i$ in $\mathcal{H}$}{
Estimate the expected value in Eq.~\eqref{eq:grad} from $\{{\bf x}^{(t)},{\bf y}^{(t)}\}$\\
$\frac{\partial(\ell(\Theta,\mathcal{D}))}{\partial{\theta_i^{(t)}}}$ = Gradient computed using Eq.~\eqref{eq:grad}\\
Update $\theta_i^{(t)}$ $=$ $\theta_i^{(t-1)}+\eta*\frac{\partial(\ell(\Theta,\mathcal{D}))}{\partial{\theta_i^{(t)}}}$\\
}
}
return $\{\theta_i^{(T)}\}_{i=1}^k$
}
\end{algorithm}
\normalsize

\subsection{Abductive Inference}

Abduction in logical reasoning is typically useful in finding the most likely explanation for an observation. In a probabilistic model, this is equivalent to the {\em Max a Posteriori} (MAP) reasoning task. Specifically, given some observed evidence, the MAP solution finds an explanation that maximizes the joint probability over all variables in the model.
%To evaluate a captioning model, we {\em abductively } infer the most likely  predicates from a ground truth caption that can explain the generated caption. 
%In the case of PGMs, this is formulated as the {\em Maximum a Posteriori} (MAP) reasoning task. Specifically, in MAP inference given observed evidence, we find an assignment to all the remaining variables in the model that has maximum probability. 
To formalize this, let ${\bf Y}^*[m]$ be the ground predicates from a human-written caption for image $m$ and ${\bf Y}[m]$ be the ground predicates generated by a model for the same image (going forward, we drop the subscript $m$ for readability). We want to infer if ${\bf Y}^*$ is the most likely explanation for the model to generate ${\bf Y}$. 

Let ${\bf M}$ represent all the variables in the model and ${\bf Y}^*$ $\subseteq$ ${\bf M}$, ${\bf Y}$ $\subseteq$ ${\bf M}$. Given an assignment to ${\bf Y}$ and the visual features from the image ${\bf X}$, the MAP predicates are $\hat{\bf M}$ $=$ ${\bf M}\setminus {\bf Y},{\bf X}$. The MAP inference task is to find an assignment to $\hat{\bf M}$ that maximizes the un-normalized log probability as follows.
%That is, we compute the most likely assignment to ${\bf y}^*$
%which can be formulated in the form of a MAP inference query that maximizes the unnormalized log probability as follows.
% In a purely logical representation,  entailment. 
%Specifically, we want to ``fill in'' the most likely assignments to {\em ground truth} predicates based on the caption generated by the model. 
%Let ${\bf y}^*[m]$ be the ground predicates in a human-written caption for image $m$ and let ${\bf X}^*[m]$ be the corresponding real-valued terms. The inference task is to estimate the conditional distribution $P_{\theta}({\bf y}^*[m]|{\bf X}^*[m])$. However, note that in a typical use case, we would want to perform inference on ${\bf y}[m]$, i.e. ground predicates from an AI generated caption. To do this, we consider the joint distribution over ${\bf y}^*[m]$, ${\bf y}[m]$ and compute a Rao-Blackwellized estimator by collapsing the variables ${\bf y}[m]$, i.e., $\sum_{y[m]\in{\bf y}[m]}P_{\theta}({\bf y}^*[m],y[m]|{\bf X}^*[m],{\bf X}[m])$. To perform exact inference is infeasible and therefore, we perform approximate Max a posteriori (MAP) inference to estimate the mode of the distribution with the following objective.
\begin{align}
\label{eq:mobj}
 \arg\max_{\hat{\bf m}\in \Omega(\hat{\bf M})}\exp\left(\sum_i \theta_is_i(\hat{\bf m},{\bf Y},{\bf X})\right)
\end{align}
%where ${\bf X}$ are the real-valued terms computed from the generated caption and $s_i(y^*|{\bf X})$ $=$ $\mathcal{C}_i(y^*[m]|{\bf X}[m])$ $+$ $\mathcal{I}_i(y^*[m]|{\bf X}[m])$.
where $\Omega(\hat{\bf M})$ is the set of all possible assignments to $\hat{\bf M}$ and $s_i(\cdot)$ is the value $\mathcal{C}_i(\cdot)+\mathcal{I}_i(\cdot)$. 

\noindent{\bf Semantic Equivalence Constraints.} To encourage the MAP solution to take into account semantic equivalences between concepts in ${\bf Y}^*$ and ${\bf Y}$, we add pairwise {\em virtual evidence} (or soft evidence) between the predicates in ${\bf Y}^*$ and ${\bf Y}$. The concept of virtual evidence was first introduced by Pearl~\cite{pearl88} to handle uncertainty in observations. Further, this has been used in several approaches since such as semi-supervised learning~\cite{kingma2014semi} and grounded learning~\cite{parikh2015grounded} as a form of indirect supervision (e.g. label preferences). In general, the idea of virtual evidence is to introduce a potential function over the evidence that encodes its uncertainty. In our case, we add the evidence based on semantic equivalence between the subject and object mentions in the predicates $Y^*\in{\bf Y}^*$ and $Y\in{\bf Y}$ respectively. This is similar to the approach used in~\cite{beltagy&al14} to add soft distributional semantics constraints to infer text entailment. Specifically, we use a standard word embedding model to compute the similarity scores between the subject pairs in $(Y^*,Y)$ and the object pairs in $(Y^*,Y)$. We consider the semantic equivalence score $S(Y^*,Y)$ as the minimum of the two similarity scores. We add the virtual evidence $(Y^*\Leftrightarrow Y)$ with value equal to $S(Y^*,Y)$.

\noindent{\bf MILP Encoding.} Computing the exact MAP solution that optimizes Eq.~\eqref{eq:mobj} is known to be NP-hard~\cite{koller&friedman09}. Therefore, we compute the approximate MAP solution with the help of a Mixed Integer Linear Programming (MILP) solver. The encoding that we use is as follows. Each symbolic feature is represented as a binary variable and a continuous feature is encoded as a real-valued variable. We introduce auxiliary variables for each formula and also for the soft evidence. Specifically, for a formula $f$ (or soft evidence) with weight $\theta$, we introduce a binary auxiliary variable $a$ with objective $\theta$ and add a hard constraint $f\Leftrightarrow a$ which is encoded as linear constraints in the MILP. In our experiments, we use a state-of-the-art solver namely, Gurobi~\cite{gurobi} to compute the MAP solution efficiently. Larger MAP objective values indicate that the explanation of the evidence (visual features of the image and ground predicates in human-written captions) using the generated caption is more likely in the HMLN distribution. Algorithm \ref{alg:mapinference} summarizes our MAP inference approach.

\begin{algorithm}[!t]{
\small
\linesnumbered
\caption{Abductive Inference}\label{alg:mapinference}
\KwIn{Parameterized HMLN with structure $\mathcal{H}$ and weights $\Theta$, ground predicates from reference captions ${\bf Y}^*$, ground predicates from generated caption ${\bf Y}$}
\KwOut{Assignment to MAP predicates $\hat{\bf M}$, MAP Objective Value $m^*$}
\tcp{MILP Encoding}
\For{each formula $f$ with parameter $\theta$ in $\mathcal{H}$}
{
$X$ $=$ Continuous terms in $f$\\
$Y$ $=$ Discrete Terms in $f$\\
Add $X,Y$ to MILP model\\
Add hard constraint $f\Leftrightarrow a$, where $a$ is an auxiliary variable with objective $\theta$\\
}
\tcp{Soft Evidence}
\For{$(Y,Y^*)$, where $Y\in{\bf Y}$, $Y^*\in{\bf Y}^*$}{
        $S$ $=$ Similarity between subjects in $Y,Y^*$\\
        $O$ $=$ Similarity between objects in $Y,Y^*$\\
        Add hard constraint $(Y\Leftrightarrow Y^*)\Leftrightarrow \hat{a}$, where $\hat{a}$ is an auxiliary variable with objective $\min(S,O)$\\
    }
$\hat{\bf m}$, $m^*$ = Assignment and objective value solving Eq.~\eqref{eq:mobj} using a MILP solver\\
return $\hat{\bf m}$, $m^*$\\
}
\end{algorithm}
\normalsize

\subsection{Back-Tracing Captions to Training Examples}

The MAP inference approach proposed in the prior section requires reference (human written) captions for test images. Therefore, for large-scale datasets, this becomes harder to scale. We next develop an approach where we do not use reference captions but instead explain the bias in the generated caption through examples from the training data. First, we formalize bias similar to its formalization in a generative model~\cite{grover2019bias}. Specifically, given the true distribution $P(\cdot)$, suppose the model generates a sample ${\bf y}$ from $P_\phi(\cdot)$, we can quantify the bias in the generative distribution with the following ratio.
\begin{equation}
\label{eq:dratio}
    w({\bf y})=\frac{P({\bf y})}{P_\phi({\bf y})}
\end{equation}
%Next, we propose a sampling-based inference method where we do not use reference captions but instead sample the training data to explain {\em learning bias} of the model. Our approach is based on quantifying bias in generative models using importance weighting~\cite{grover2019bias}.
%n generative models, one can quantify {\em bias} in the generated samples using 
%Rao-Blackwellized estimate
%For models that generate samples, we can quantify {\em bias} in the generated samples using importance weighting~\cite{grover2019bias}. 

%Specifically, suppose samples ${\bf x}$ are generated under a distribution $P_{\phi}$, the bias w.r.t the true distribution $P$ is quantified as the ratio $\frac{P({\bf x})}{P_\phi({\bf x})}$. 
The ratio in Eq.~\eqref{eq:dratio} is also called as the {\em importance weight} of a sample drawn from $P_\phi(\cdot)$ when the true distribution is $P(\cdot)$. This is commonly used in importance sampling~\cite{liu01} where samples drawn from a proposal distribution $P_\phi(\cdot)$ are used to estimate expectations over $P(\cdot)$. The closer $P_\phi(\cdot)$ is to $P(\cdot)$, samples from $P_\phi(\cdot)$ have smaller bias. 

%Typically, the ratio is intractable to compute exactly. However, given samples from $P_{\phi}(\cdot)$, we can estimate the importance weight based on these samples. 

%Based on this principle, our approach is as follows. We sample from the HMLN distribution conditioned on the observed caption generated by the model. We then weight the sample by computing the unnormalized probability for the caption based on instances in the training data. Finally, we estimate the marginal densities for training instances based on the weighted samples.

% We estimate the importance weight by sampling from the prior distribution (learned from training data) and the conditional distribution after observing the generated caption. 

%We quantify the contribution of training samples to the importance weight as a measure of learning bias in the model that has generated the caption.

We apply importance weighting in our approach as follows. Let ${\bf Y}$ be the ground predicates extracted from the generated caption and let $P({\bf Y}|{\bf X})$ represent the distribution over the ground predicates conditioned on the visual features ${\bf X}$ of the test image. Let $\hat{\bf X}$ be the visual features from a training image. Given an sample ${\bf y}$ from $P({\bf Y}|{\bf X})$, its importance weight is defined as follows.

%Let ${\bf Y}$ be the ground predicates extracted from a caption generated by a model.
%Let $\mathcal{X}$ be an instance in the training data such that $\forall Y\in{\bf Y}$, $\exists X\in\mathcal{X}$ where $Y\Leftrightarrow X$. Similar to virtual evidence in abductive inference, here, we compute the semantic equivalence score between $Y,X$ using Spacy embeddings. We consider $Y\Leftrightarrow X$ iff their semantic equivalence score is above a threshold. 

%Let $P({\bf Y}|\hat{\bf X})$ be the HMLN distribution conditioned on real-valued terms that are computed using $\mathcal{X}$ and $P({\bf Y}|{\bf X})$ be the HMLN distribution conditioned on the real-valued terms ${\bf X}$ computed using the test instance. We define the importance weight as follows.
\begin{equation}
    \label{eq:impwt}
    w({\bf y})=\frac{P({\bf y}|\hat{\bf X})}{P({\bf y}|{\bf X})}  
\end{equation}
%Specifically, in the prior distribution we condition over similarities between the generated caption and the training images while in the posterior distribution we condition over similarities between the generated caption and the test images. 

%Clearly, $W({\bf Y})$ $=$ 1 implies that the model required no learning bias, i.e., the model did not need to make additional assumptions to generate the observed caption. On the other hand, a small (or large) $w({\bf Y})$ indicates that the model has a stronger learning bias, i.e., the difference between the distribution from which the caption was sampled for the test image is different from the training distribution and therefore the model needed to add inference-time learning bias to generate the observed caption. Note that for the sample to be {\em properly weighted}, whenever the numerator is positive (non-zero) denominator must be positive. In our case, we can show that if there are no {\em hard constraints} in the HMLN (formulas with weight $\infty$), then since there is no determinism or 0 probability worlds, the importance weight $w({\bf Y})$ is properly weighted.

\noindent{\bf Marginal Density Estimation.} 
We estimate the marginal densities over training examples using the importance weights. However, the number of training examples is typically very large. Further, most of the training examples will typically have little relevance to a test example. Therefore, to scale-up we first select a small number of training examples that are {\em contextually relevant} to the test example. To do this, we only consider a training example $\mathcal{X}\in\mathcal{D}$ that is grounded in the same context as the test image. Specifically, we consider $\mathcal{X}$ to be contextually relevant to a test example $\mathcal{Y}$ iff for each ground predicate $X$ from the caption for $\mathcal{X}$, there exists a ground predicate $Y$ in the generated caption for $\mathcal{Y}$ such that $S(X,Y)>C$, where $S(X,Y)$ is the score measuring semantic equivalence between $X$ and $Y$, and $C$ is a constant.
%a semantic equivalence score larger than a th to at least one ground predicate in the generated caption for $\mathcal{I}$.
%We compute the similarity between predicates by comparing the word embeddings of their subject mentions and their object mentions respectively.

Let $\mathcal{X}$ be a training image that is contextually relevant to the test image and let ${\bf Y}$ be ground predicates in the generated caption. The density $P(\mathcal{X}|{\bf Y})$ is the likelihood that a model that learns from $\mathcal{X}$ has generated ${\bf Y}$. We assume a {\em mean-field} distribution over all related training examples and therefore in factorized form, we can write its joint distribution as follows.
\begin{equation}
\label{eq:joint}
P(\mathcal{X}_1\ldots\mathcal{X}_n|{\bf Y})=\prod_{i=1}^nP(\mathcal{X}_i|{\bf Y})    
\end{equation}
Given a sample ${\bf y}$ over the ground predicates, we define the following indicator function. $\mathbb{I}_{\mathcal{X}}[{\bf y}]$ $=$ $1$ iff ${\bf y}$ contains an assignment to predicates such that for each predicate that exists in $\mathcal{X}$, there is a semantically equivalent predicate that has an assignment equal to 1 in in ${\bf y}$. We can compute the expected value of the indicator function using importance weights as follows.
%whenever the iff given a sample, i.e., assignments over all ground predicates, there exists a predicate that is assignment equivalences in $\mathcal{X}$. 
%$f$
%Note that we can estimate any expectation $f(y)$ from the importance weighted samples as follows.
%We define the expected learning bias of a model that generates a caption with ground predicates ${\bf Y}$ as follows.
$$\mathbb{E}_{{\bf y} \sim P({\bf Y}|\hat{\bf X})}\left[\mathbb{I}_{\mathcal{X}}[{\bf y}]\right]=\mathbb{E}_{{\bf y} \sim P({\bf Y}|{\bf X})}\left[\frac{P({\bf y}|\hat{\bf X})}{P({\bf y}|{\bf X})}\mathbb{I}_{\mathcal{X}}[{\bf y}]\right]$$
%where $y$ is an assignment to ${\bf Y}$ and $f(y)$ is a function over $y$. Let $f(y)$ denote an indicator variable which is true if 
%denote the value of a formula $f$ given the assignment $y$. Thus, Eq.~\ref{} is simply an estimation of the expected value of a formula w.r.t to the posterior distribution. 
Since computing the exact expectation is intractable, given $T$ samples ${\bf y}_1\ldots {\bf y}_T$, we can obtain a Monte-Carlo estimate of the expected value as follows.
%given importance weights $\{w(y_i)\}_{i=1}^T$.
\begin{equation}
\label{eq:est1}
  \mathbb{E}_{{\bf y} \sim P({\bf Y}|\hat{\bf X})}\left[\mathbb{I}_{\mathcal{X}}[{\bf y}]\right] \approx \frac{1}{T}\sum_{i=1}^Tw({\bf y}_i)\mathbb{I}_{\mathcal{X}}[{\bf y}_i]  
\end{equation}

Thus, using samples drawn from $P({\bf Y}|\hat{\bf X})$, we can estimate the expectation in Eq.~\eqref{eq:est1}. From the results in~\cite{liu01}, we can show that the estimator is unbiased, i.e., we can show that as $T\rightarrow\infty$, the estimates converge to the true expectation. However, a technical challenge is that it is hard to estimate the true importance weights. Specifically, note that the distributions in the numerator and denominator in Eq.~\eqref{eq:impwt} have different normalization constants. Therefore, to estimate the exact weights, we need to estimate the normalization constants of both the distributions which is computationally intractable. Therefore, we instead use an approximate estimator where it is tractable to estimate the importance weights. Specifically, we compute the ratio of the un-normalized probabilities to compute the weights. In this case, the computation is efficient since we do not need the normalization constants. Thus, we have a modified weight as follows.

\begin{equation}
    \label{eq:approxwt}
    w({\bf y})\propto\frac{P({\bf y}|\hat{\bf X})}{P({\bf y}|{\bf X})}  
\end{equation}

%We can compute the sample weight easily since it is the ratio of unnormalized probabilities. However, note that since the normalization constants in the numerator and the denominator of the importance weight are different, the unbiased estimator in Eq.~\eqref{eq:est1} cannot be used. 
We can show that using the weights in Eq.~\eqref{eq:approxwt}, the estimated expectation is {\em asymptotically unbiased}~\cite{liu01}, i.e., the bias goes to 0 as $T\rightarrow\infty$.
%the expected values computed from the importance weights are not unbiased
A second practical difficulty is that some weights may get too large and this tends to yield poor estimates. To ensure that weights do not get too large, we use an approach called {\em clipping}  which thresholds the maximum weights. This ensures that a few weights do not dominate in the estimation of the expectation. The modified estimator combining the un-normalized weights and clipping is as follows.
\begin{equation}
\label{eq:est2}
   P(\mathcal{X}|{\bf Y})\approx \frac{\sum_{i=1}^T\max(w({\bf y}_i,1))\mathbb{I}_{\mathcal{X}}[{\bf y}_i]}{\sum_{i=1}^T\max(w({\bf y}_i,1))}
\end{equation}

As $T\rightarrow\infty$, the estimate of the expectation computed in Eq.~\eqref{eq:est2} converges to true marginal density $P(\mathcal{X}|{\bf Y})$. To implement this, we use Gibbs sampling. Specifically, we perform Gibbs sampling to draw samples from $P({\bf y}|{\bf X})$ and weight each sample by computing the unnormalized ratio in Eq.~\eqref{eq:approxwt}.
%Thus, the final estimate of the expectation which is also our estimate of the marginal over $\mathcal{X}$ is as follows. 
%Note that we only update the estimates for the marginal probability once the Gibbs sampler has {\em mixed}, i.e., the MCMC sampler generates samples from the target distribution. To do this, the typical approach is to ignore the first few samples (called the {\em burn-in} period) in the estimation. 
Strictly speaking, for the estimator in Eq.~\eqref{eq:est2}, we would want the samples drawn from $P({\bf y}|{\bf X})$ to be independent of each other. However, Gibbs sampling generates dependent samples. That is, the sample generated in an iteration depends upon previous samples. To address this, we use a technique called {\em thinning}, where we only consider samples for estimating the weight after every $k$ Gibbs iterations. This reduces the dependence among samples. 
%Since the estimator is asymptotically unbiased, as $T\rightarrow$ $\infty$ the expected value computed in Eq.~\eqref{eq:est2} converges to $P(\mathcal{X}|{\bf Y})$. 
%$\underset{\bar{}}{\mathcal{X}}_{\bf Y}$ 
%Further, contrastive explanations~\cite{} help a user understand the training examples that have similar context as the test example implications have varying effects on a model's learning.

\noindent{\bf Contrastive Samples.} 
Humans often compare contrastive examples to understand concepts~\cite{miller2019explanation}. Therefore, in our case, we use the distribution in Eq.~\eqref{eq:joint} to sample instances that have contrasting influence on the generated caption.
Specifically, the minimal factor in the distribution, $\mathcal{X}^-$ $=$ $\arg\min_{i=1}^nP(\mathcal{X}_i|{\bf Y})$ denotes the training instance that is the least likely to generate ${\bf Y}$. The maximal factor $\mathcal{X}^+$ $=$ $\arg\max_{i=1}^nP(\mathcal{X}_i|{\bf Y})$ denotes the most likely instance to generate ${\bf Y}$. Algorithm \ref{alg:marginference} summarizes our approach.

%Therefore, the model that generates ${\bf Y}$ should typically use $\underset{\bar{}}{\mathcal{X}}_{\bf Y}$ as a positive example and $\bar{\mathcal{X}}_{\bf Y}$ as a negative example in generating the caption.

\eat{
\begin{algorithm}[!t]{
\small
\linesnumbered
\caption{Back-Tracing}\label{alg:marginference}
\KwIn{Parameterized HMLN with structure $\mathcal{H}$ and weights $\Theta$, ground predicates from generated caption ${\bf Y}$, relevant training examples $\{\mathcal{X}_i\}_{i=1}^n$}
\KwOut{Explanation for ${\bf Y}$}
\For{each $\mathcal{X}_i$}{
    \tcp{Gibbs Sampling}
    ${\bf y}^{(0)}$ $=$ Initialize the assignments to all discrete ground predicates in $\mathcal{H}$\\
    ${\bf X}$ $=$ Continuous terms computed from test image CLIP embedding\\
    \For{$t$ $=$ 1 to $T$}
    {
        ${\bf y}^{(t)}$ $=$ Gibbs Sample from the posterior distribution\\
        \If{burn-in completed}{
            $w({\bf y}^{(t)})$ $=$ ratio computed using Eq.~\eqref{eq:impwt}\\
            Update marginal estimate for $\hat{P}(\mathcal{X}_i)$ using the estimator in Eq.~\eqref{eq:est2}\\
        }
    }
}
return $\arg\max_{i=1}^n\hat{P}(\mathcal{X}_i)$, $\arg\min_{i=1}^n\hat{P}(\mathcal{X}_i)$\\
}
\end{algorithm}
\normalsize
}

\begin{algorithm}[!t]{
\small
\linesnumbered
\caption{Marginal Inference}\label{alg:marginference}
\KwIn{Ground predicates from generated caption ${\bf Y}$, relevant training examples $\{\mathcal{X}_i\}_{i=1}^n$}
\KwOut{Contrastive training samples for ${\bf Y}$}

\For{each Gibbs sample ${\bf y}^{(t)}$ from the target distribution $P({\bf Y}|{\bf X})$}
{
    Compute the un-normalized sample weight $w({\bf y}^{(t)})$ using Eq.~\eqref{eq:approxwt}\\
    \For{each $\mathcal{X}_i$}{
    Update its marginal density using the estimator in Eq.~\eqref{eq:est2}\\
    }
}
return $\arg\max_{i=1}^n\hat{P}(\mathcal{X}_i)$, $\arg\min_{i=1}^n\hat{P}(\mathcal{X}_i)$\\
}
\end{algorithm}
%Thus, if a model generates ${\bf Y}$, we can interpret the model's learning bias 

%Note that since the normalization constants in the numerator and the denominator of the importance weight are different, the expected values computed from the importance weights are not unbiased but {\em asymptotically unbiased}. That is as $T\rightarrow$ $\infty$, the bias reduces to 0. In this case, the expected value converges to $P(\mathcal{X}|{\bf Y})$.

\section{Experiments}

%Using our proposed methods, we evaluate i) classical captioning models (SGAE~\cite{yang2019auto}), ii) transformer-based models (AoA~\cite{huang2019attention}, XLAN~\cite{pan2020x}, M2~\cite{cornia2020meshed}) and iii) Visual Large Language Models (VLLMs) (Blip2~\cite{li2023blip}). 

%In each case, we used their publicly available pretrained models for the evaluation. We used the MSCOCO image captioning benchmark dataset with Karpathy's train, test, validation split~\cite{karpathy2015deep}. 

%The training data consists of 113K images whereas the validation set and test set consists of 5K and 5K images respectively. The number of captions per image is equal to 5. 

\subsection{Implementation}

To implement our approach, we used the textual scene graph parser~\cite{schuster2015generating} to obtain ground predicates from the text of captions. We used the pretrained CLIP model to embed the image and text to encode real-valued terms in the HMLN. For weight learning in the HMLN, we set the learning rate to 0.01. To implement abductive inference, we used the MILP solver in Gurobi to obtain the approximate MAP solution. For determining semantic equivalence between ground predicates, we used SpaCy word embedding similarity scores~\cite{honnibal2020spacy}. For the Gibbs sampler, we used a {\em burn-in} of 500 samples. After this, we estimated the marginal probabilities using the clipped estimator with an interval of 10 iterations between samples to minimize dependency across samples. We performed our experiments on a 62.5 GiB RAM, 64-bit Intel® Core™ i9-10885H CPU @ 2.40GHz × 16 processor with a NVIDIA Quadro GPU with 16GB RAM. Our code is available at the following link. \textcolor{blue}{\em https://github.com/Monikshah/caption-explanation-hmln}
\eat{
\begin{table}[!t]
    \centering
    \resizebox{0.45\textwidth}{!}{
    \tabcolsep=0.15cm
    \begin{tabular}{|c|c|c|c|c|c|}
        \hline
        \textbf{..} & \textbf{SGAE} & \textbf{AOA} & \textbf{M2} & \textbf{XLAN} & \textbf{BLIP2} \\
        \hline
        \hline
        \textbf{$Bleu_1$}	& 0.810 & 0.805 & 0.807 & 0.808 & 0.781 \\
        \textbf{$Bleu_2$}	& 0.656 & 0.652 & 0.653 & 0.657 & 0.627 \\
        \textbf{$Bleu_3$}   & 0.508 & 0.509 & 0.509 &	0.515 &	0.483 \\
        \textbf{$Bleu_4$}   & 0.387 & 0.391 & 0.391 &	0.396 &	0.364 \\
        \textbf{$METEOR$}   & 0.284 & 0.290 & 0.291 &	0.295 &	0.314  \\ 
        \textbf{$ROUGE_L$}  & 0.586 & 0.589 & 0.586 &	0.591 &	0.591 \\
        \textbf{$CIDER$}    & 1.283 & 1.289 & 1.312 &	1.328 &	1.316 \\
        \textbf{$SPICE$}    & 0.217 & 0.226 & 0.226 &	0.234 &	0.252 \\
        \textbf{$MAP$}   & 0.689 & 0.682 & 0.681 & 0.825 & 0.665 \\
        \hline
    \end{tabular}
    }
    \caption{Comparing reference-based metrics.}
    \label{tab:metrics}
\end{table}
}

\subsection{Models}
We evaluate our approach on the state-of-the-art image captioning models including SGAE~\cite{yang2019auto}, AoA~\cite{huang2019attention}, XLAN~\cite{pan2020x}, M2~\cite{cornia2020meshed} and Blip2~\cite{li2023blip}. In each case, we used their publicly available models for the evaluation. Each of these models have been pretrained over datasets. Specifically, SGAE and XLAN are pretrained  using ImageNet (IN) \cite{deng2009imagenet} and Visual Genome (VG) \cite{krishna2017visual}, AoA and M2 are pretrained with VG. BLIP2 is pretrained with the LLM OPT~\cite{zhang2022opt}. It is also pretrained with datasets VG, CC3M \cite{sharma2018conceptual}, CC12M \cite{changpinyo2021conceptual} and SBU \cite{ordonez2011im2text}).
All the models are fine-tuned for generating captions on the MSCOCO image captioning benchmark dataset with Karpathy's train, test, validation split. We use the same split to learn our HMLN and perform evaluation on the test split.
% We evaluate our approach on i) Classical captioning models (SGAE~\cite{yang2019auto}), ii) State-of-the-art transformer models (AoA~\cite{huang2019attention}, XLAN~\cite{pan2020x}, M2~\cite{cornia2020meshed}) and iii) State-of-the-art Visual Large Language Models (VLLMs) (Blip2~\cite{li2023blip}). In each case, we used their publicly available models for the evaluation. Each of these models have been pretrained over datasets. Specifically, SGAE and XLAN are pretrained  using ImageNet (IN) \cite{deng2009imagenet} and Visual Genome (VG) \cite{krishna2017visual}, AoA and M2 are pretrained with VG. BLIP2 is pretrained with the LLM OPT~\cite{zhang2022opt}. It is also pretrained with datasets VG, CC3M \cite{sharma2018conceptual}, CC12M \cite{changpinyo2021conceptual} and SBU \cite{ordonez2011im2text}).
% All the models are fine-tuned for generating captions on the MSCOCO image captioning benchmark dataset with Karpathy's train, test, validation split. We use the same split to learn our HMLN and perform evaluation on the test split.
%The training data consists of 113K images whereas the validation set and test set consists of 5K and 5K images respectively. 
%The number of human-written captions per image is equal to 5. 
%BLIP2: (LLM\cite{zhang2022opt, chung2024scaling}, 
\subsection{MAP Inference}

\begin{figure}
    \centering
    \includegraphics[scale=0.36]{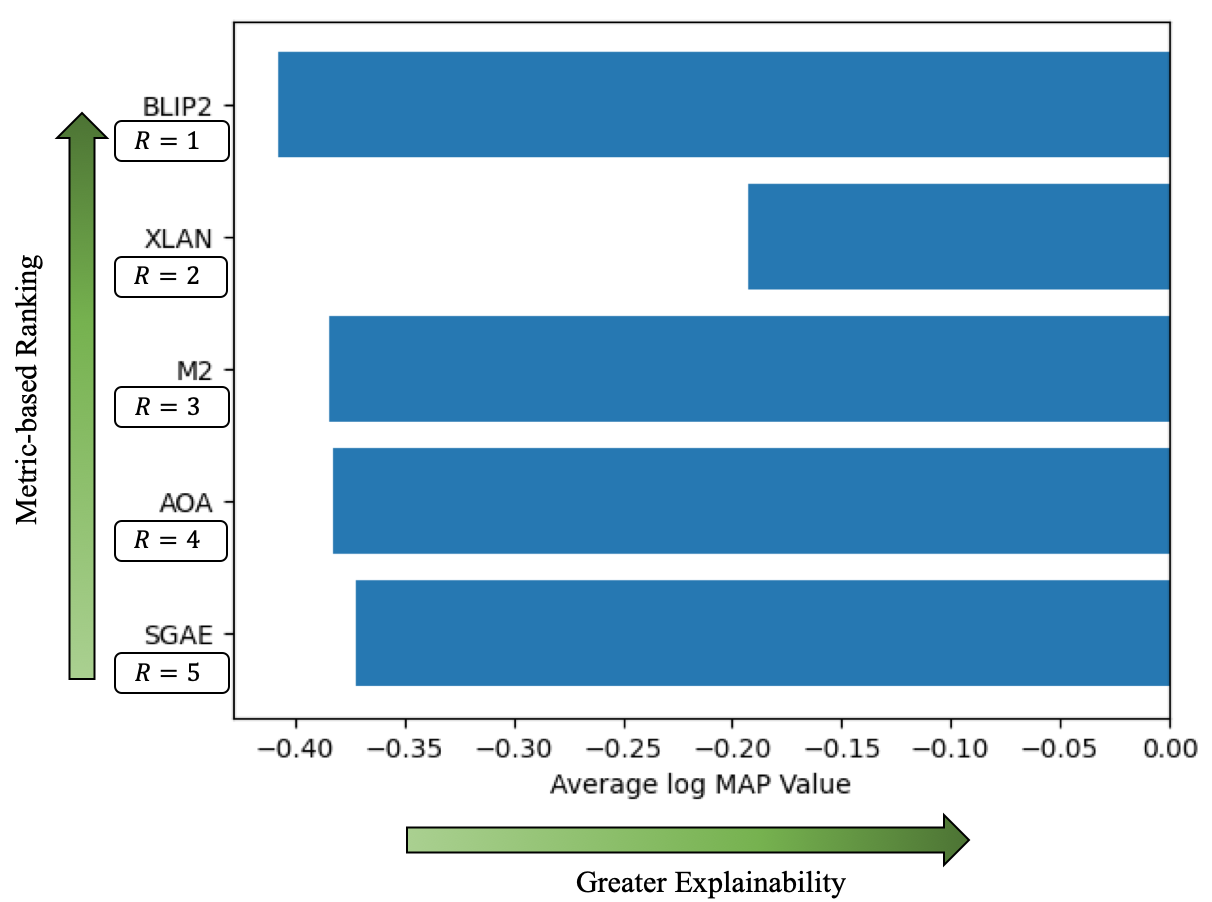}
    \caption{The x-axis shows the average log MAP objective value computed across test images. The y-axis shows the ranking of a model where the rank is computed based on the average rank across 5 standard captioning metrics ($R=1$ is the highest ranked model).}
    \label{fig:mapvalues}
\end{figure}

The average MAP objective value over captions generated for the test set is compared over all five models in Fig.~\ref{fig:mapvalues}. A larger average MAP value (smaller negative-log value) indicates that the generated captions from the model are more likely to have maximal probability in the distribution.
Note that for the VLLM model BLIP2, the LLM improves general knowledge that improves captioning ability of the model. However, the generation is harder to explain with the training distribution. This trade-off can be illustrated in our results. Specifically, we compute the average rank for a model based on 5 standard captioning metrics (BLEU4, METEOR, ROUGE, CiDER and SPICE) which indicate caption quality. The ranking is indicated in Fig.~\ref{fig:mapvalues}, where BLIP2 has the highest rank followed by XLAN. However, XLAN has a significantly higher MAP value than BLIP2. Note that the other non-VLLM models with similar pretraining as XLAN have smaller MAP values since their generated captions do not explain the visual features as accurately as the captions generated by XLAN. 

Some illustrative examples are shown in Fig.~\ref{fig:examplesmap}. VLLMs encode more general knowledge that can help it generate captions for test examples that are hard to learn from training examples. For instance, in Fig.~\ref{fig:examplesmap} (a), it is hard to infer the {\em refrigerator} in the image, however the logo {\em GE} is visible and using pretraining from LLMs, BLIP2 can make the inference that it represents a refrigerator while a non-VLLM cannot. However, the smaller MAP score shows that explaining such inference is harder for VLLMs. On the other hand, for examples Fig.~\ref{fig:examplesmap} (b), (c) the MAP values are high for both BLIP2 and the non-VLLM. However, BLIP2 includes more details within the caption (e.g. {\em basil}, {\em cactus}).
In Fig.~\ref{fig:examplesmap} (d), BLIP2's LLM-based knowledge has the opposite effect since it infers that the bottle is a {\em fire extinguisher} due to its color. In Fig.~\ref{fig:examplesmap} (e), the smaller MAP score for both the VLLM and non-VLLM model indicates that it is hard to explain the image using training data due to its unique characteristics (an unusual {\em flying kite}). Finally, in Fig.~\ref{fig:examplesmap} (f), BLIP2 can identify the banana in a novel context compared to the non-VLLM and since this explains the visual features in the image, MAP inference reflects this in its larger value.

%The trade-off between explainability and captioning ability (measured by the metric-based ranking) can be seen for the VLLM (BLIP2).
\eat{
\begin{figure}
    \centering
    \includegraphics[scale=0.6]{images/mapranks.pdf}
    \caption{Caption}
    \label{fig:mapranks}
\end{figure}
}

\eat{
\begin{figure}
\centering
    \includegraphics[width=80mm]{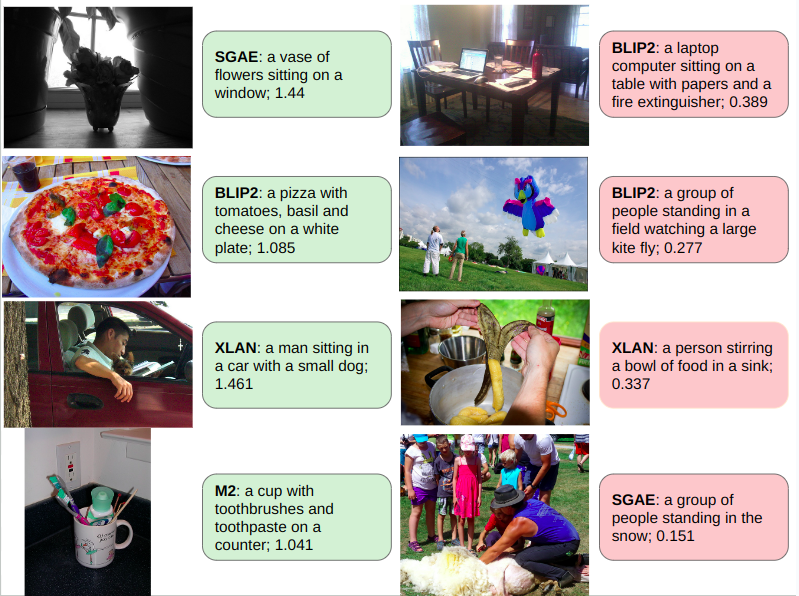}
    \caption{Examples comparing XLAN and Blip2 MAP values.}
    \label{fig:examplesmap}
\end{figure}
}
\eat{
\begin{figure}
    \centering
    \subfigure[VLLM]{\includegraphics[scale=0.35]{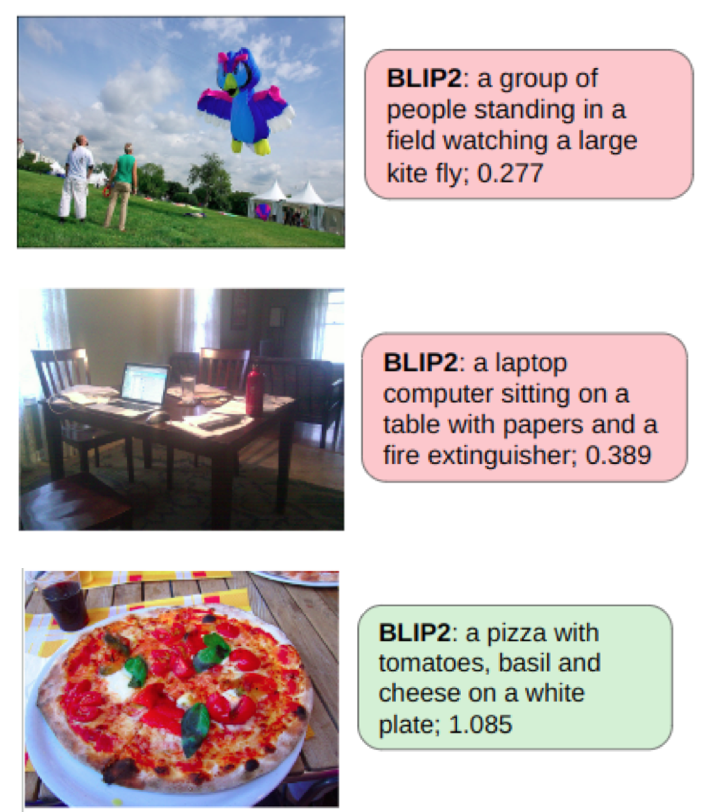}}
    \subfigure[Non-VLLM]{\includegraphics[scale=0.35]{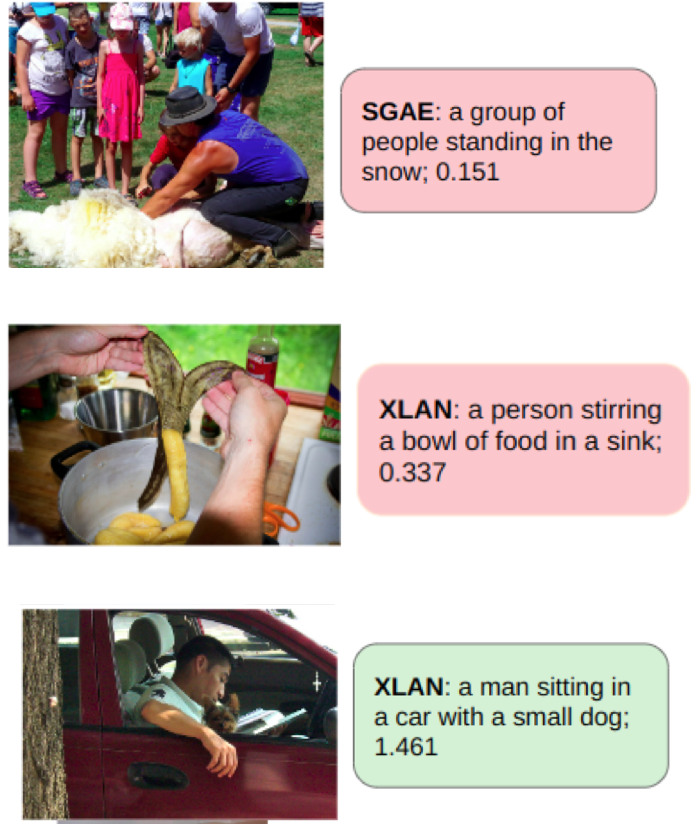}}

    \caption{Illustrative examples for MAP values of captions for BLIP2 and non-VLLM models. The box in green has the largest MAP value in both cases.}
    \label{fig:examplesmap}
\end{figure}
}

\begin{figure*}
\centering
    \subfigure[]{\includegraphics[scale=0.36]{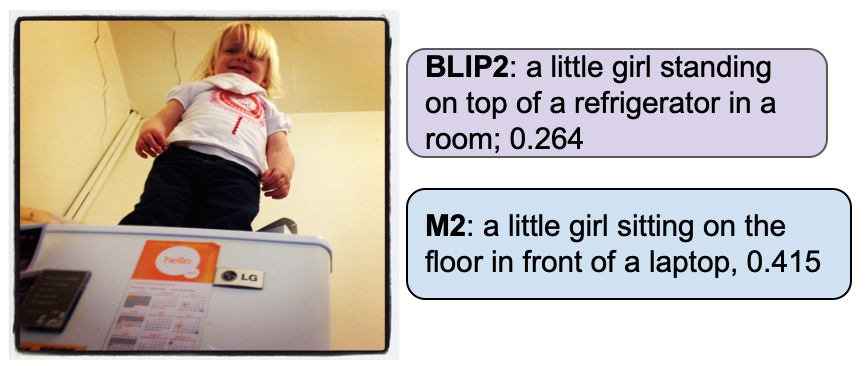}}
    \subfigure[]{\includegraphics[scale=0.36]{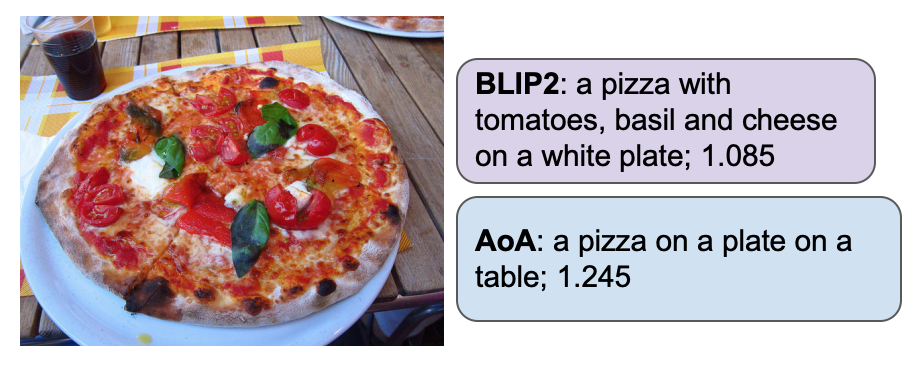}}
    \subfigure[]{\includegraphics[scale=0.36]{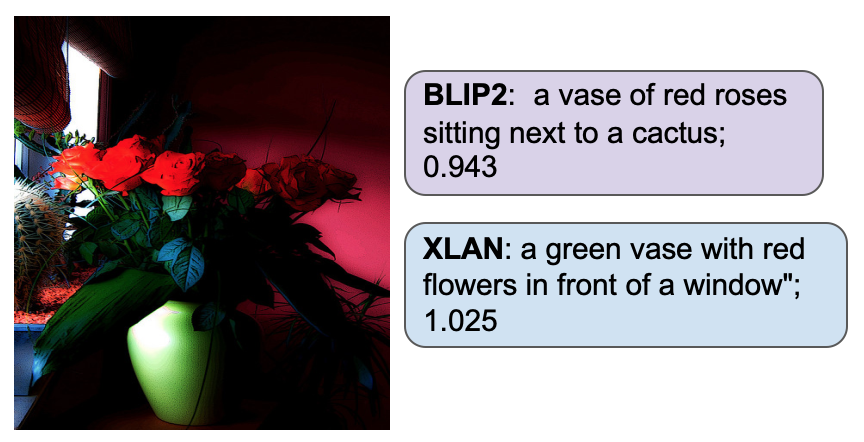}}
    \subfigure[]{\includegraphics[scale=0.36]{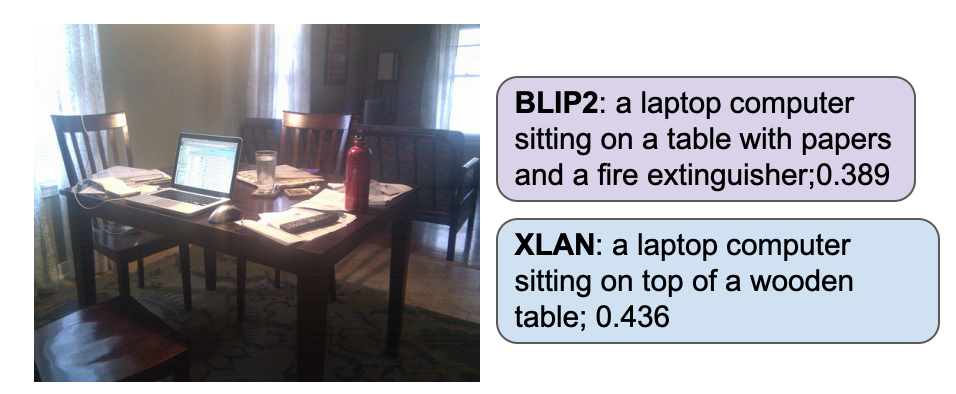}}
    \subfigure[]{\includegraphics[scale=0.36]{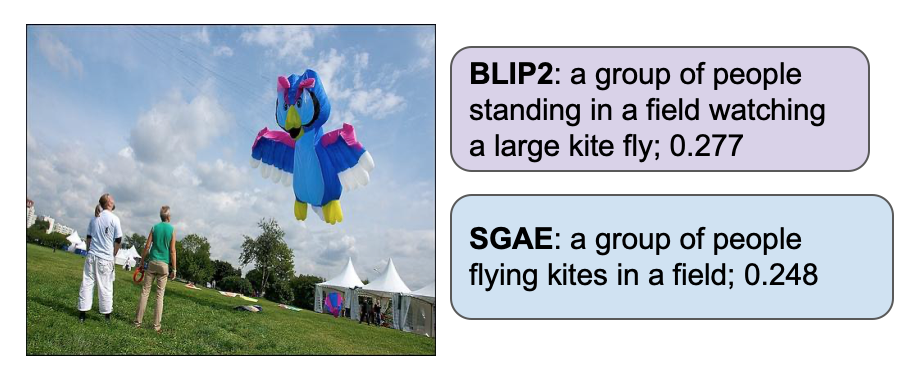}}
    \subfigure[]{\includegraphics[scale=0.36]{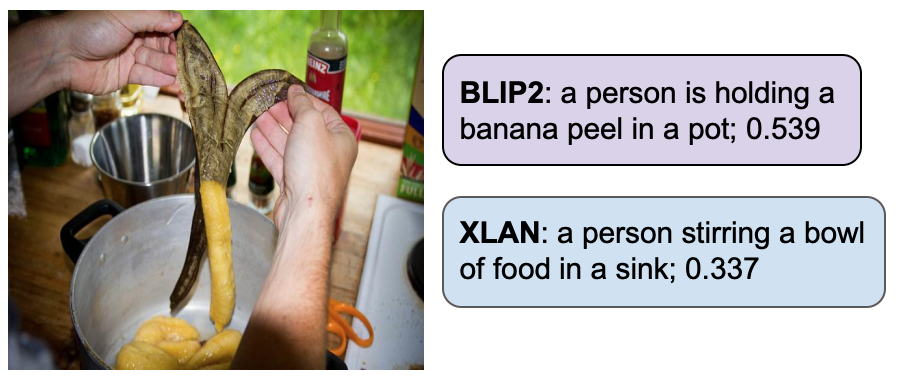}}
     \caption{\label{fig:examplesmap}Illustrative examples for MAP values of captions for BLIP2 and non-VLLM models.}
\end{figure*}

% \begin{figure}
%     \includegraphics[width=80mm]{images/turkexpl-new.png}
%     \caption{Caption}
%     \label{fig:expl}
% \end{figure}

\begin{figure}
\centering
        \includegraphics[scale=0.65]{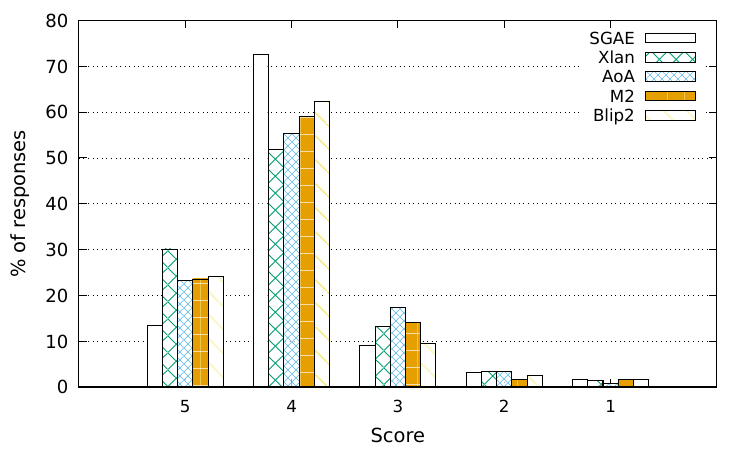}
    \caption{\label{fig:turkresults1} Comparing responses (from a Likert scale of 1-5) from AMT users across models for explaining their generated captions using training examples. Higher values indicate that the users more strongly agreed with the explanations.}
\end{figure}

\eat{
\begin{figure}
\centering
    \includegraphics[width=80mm]{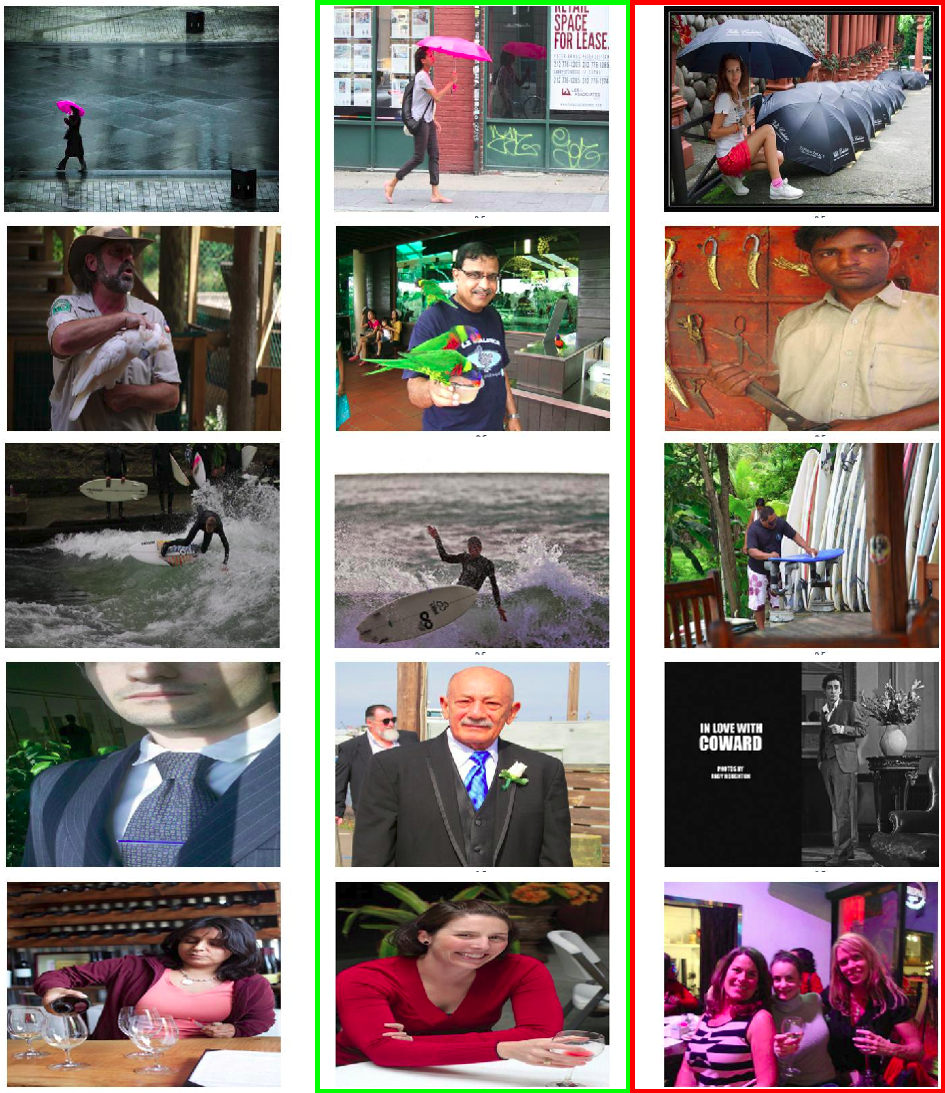}
    \caption{Examples of maximal and minimal marginals. The first image is the test image, the second one (green border) is the training instance with maximum marginal value and the third one (red border) is the training instance with minimal marginal value.}
    \label{fig:exampleamt}
\end{figure}
}

\begin{figure*}
\centering
    \subfigure[BLIP2: A woman walking down a street with a pink umbrella.]{\fbox{\includegraphics[scale=0.25]{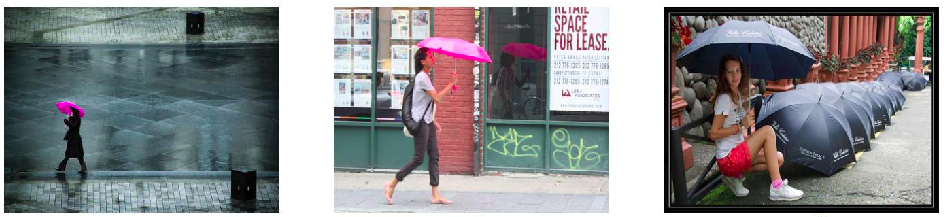}}}
    \subfigure[XLAN: A room with a table and a clock on the wall.]{\fbox{\includegraphics[scale=0.25]{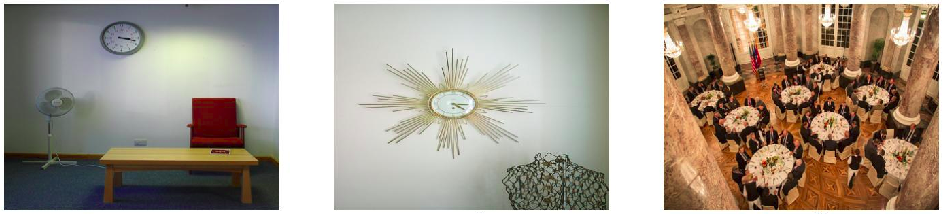}}}
    \subfigure[M2: A woman sitting at the table eating a hot dog.]{\fbox{\includegraphics[scale=0.25]{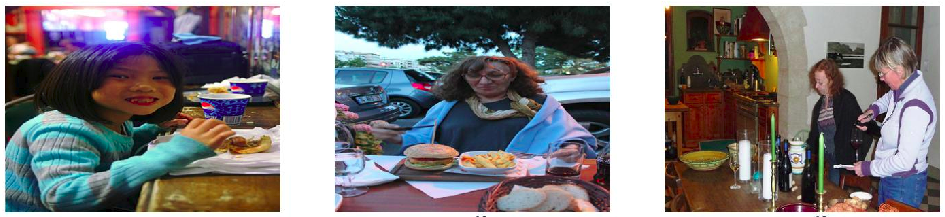}}}
    \subfigure[AOA: A bowl of apples and oranges on a table.]{\fbox{\includegraphics[scale=0.25]{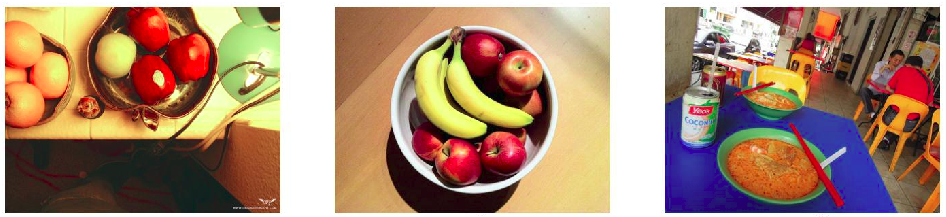}}}
     \caption{\label{fig:examplesexplain}Illustrative contrastive training examples for the generated captions. The caption shown in each case was generated for the first image, the second image is the largest probability example and the third one is the smallest probability training example.}
\end{figure*}

\eat{
\begin{figure*}
    \centering
    \begin{subfigure}{0.3\linewidth}
        \includegraphics[scale=0.2]{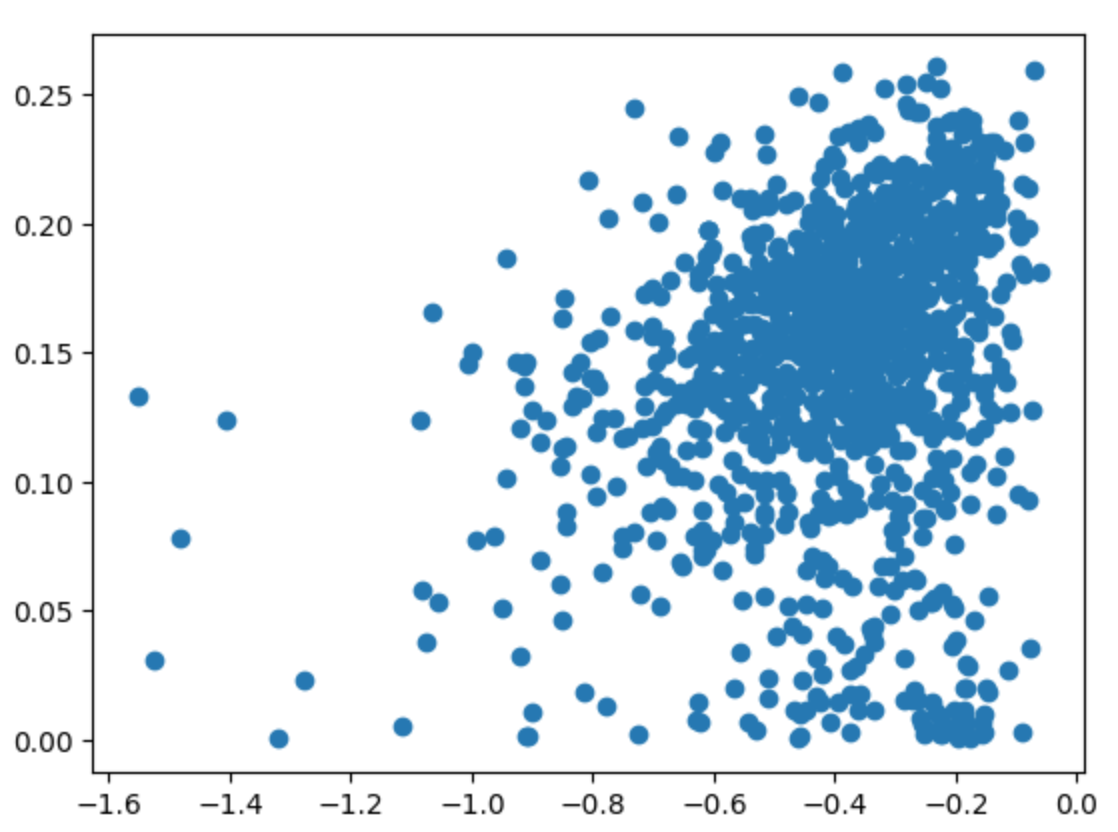}
        \caption{SGAE}
        \label{fig:sg}
    \end{subfigure} %
    \begin{subfigure}{0.3\linewidth}
        \includegraphics[scale=0.2]{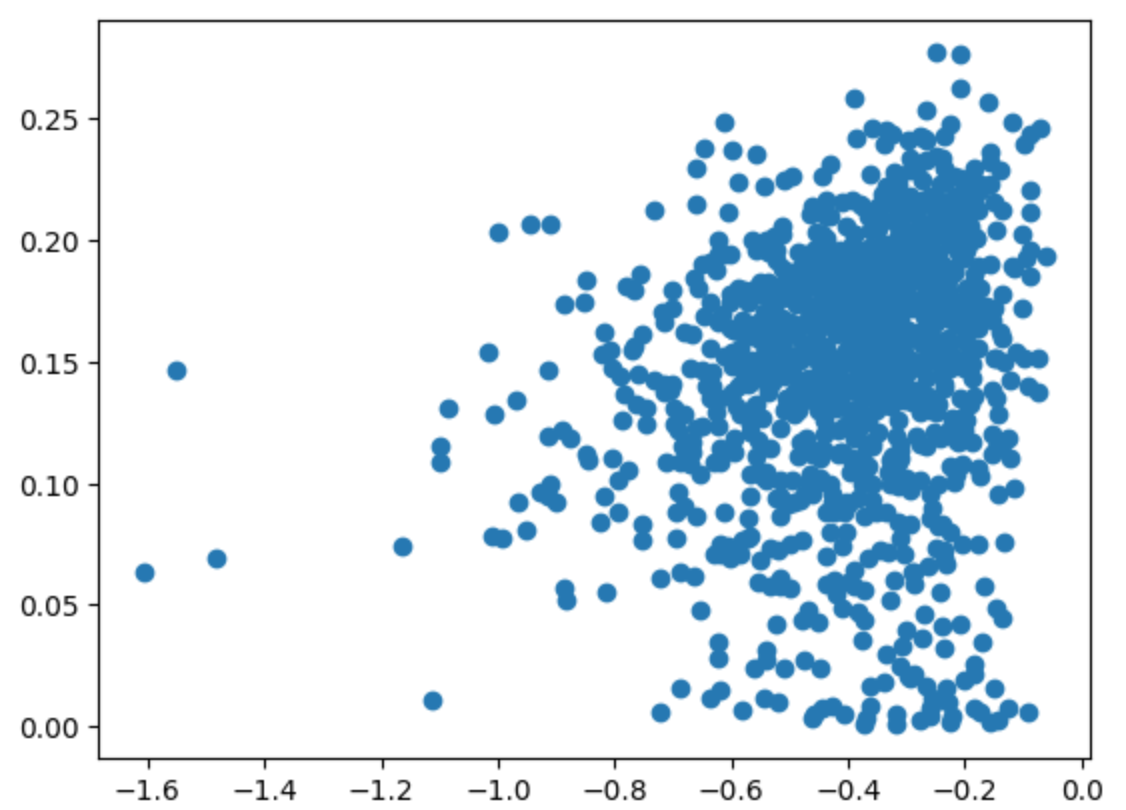}
        \caption{AoA}
        \label{fig:aoa}
    \end{subfigure} %
    \begin{subfigure}{0.3\linewidth}
        \includegraphics[scale=0.2]{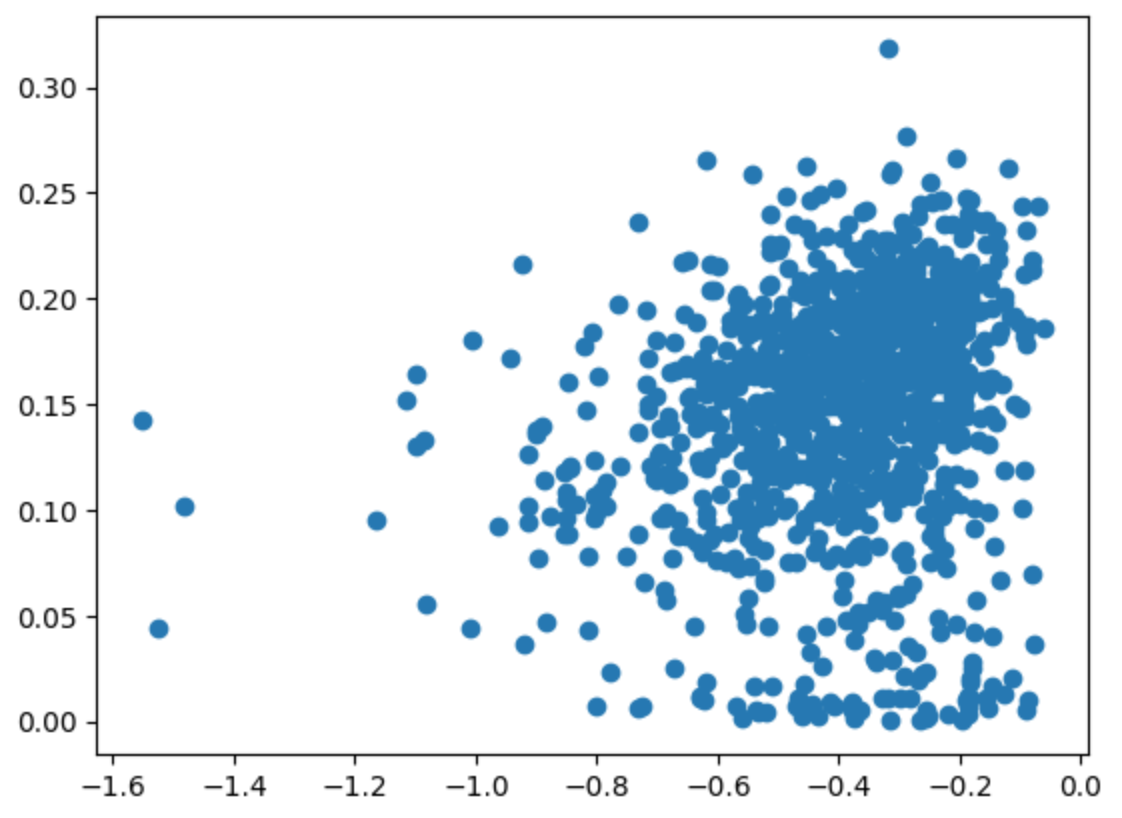}
        \caption{M2}
        \label{fig:sg}
    \end{subfigure} %
    \begin{subfigure}{0.3\linewidth}
        \includegraphics[scale=0.2]{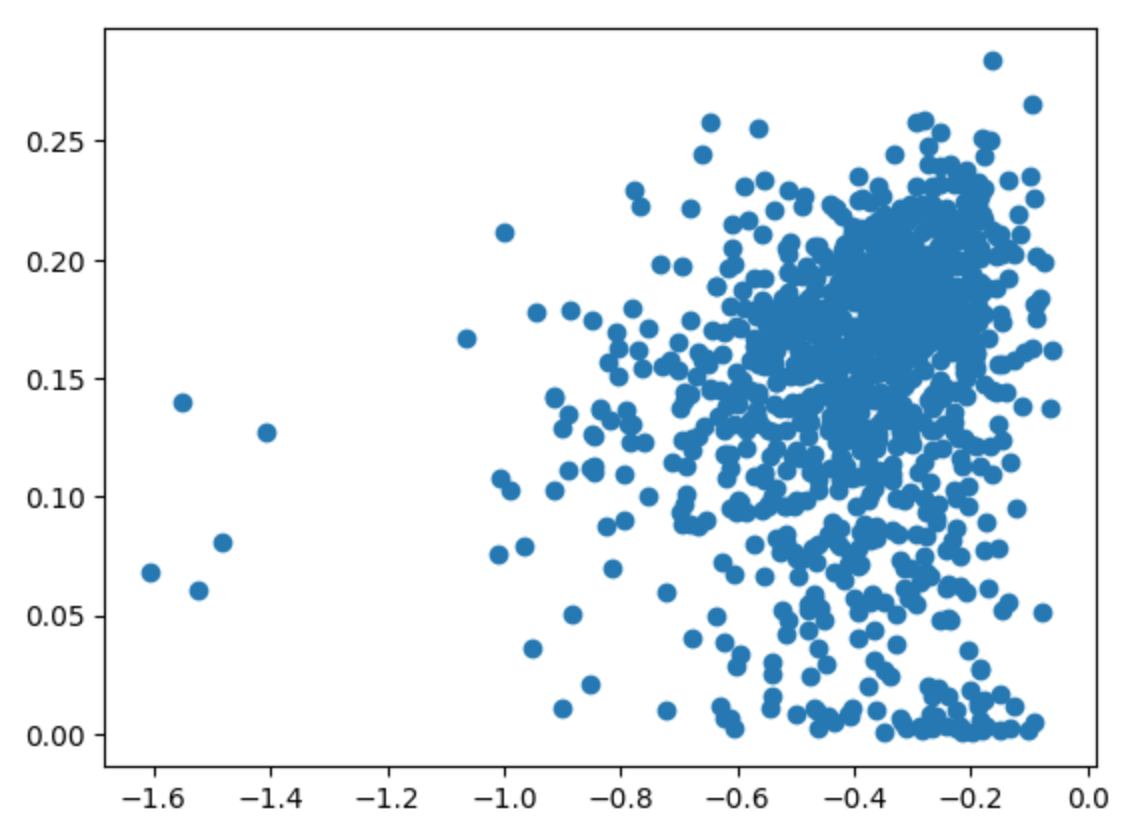}
        \caption{XLAN}
        \label{fig:sg}
    \end{subfigure} %
    \begin{subfigure}{0.3\linewidth}
        \includegraphics[scale=0.2]{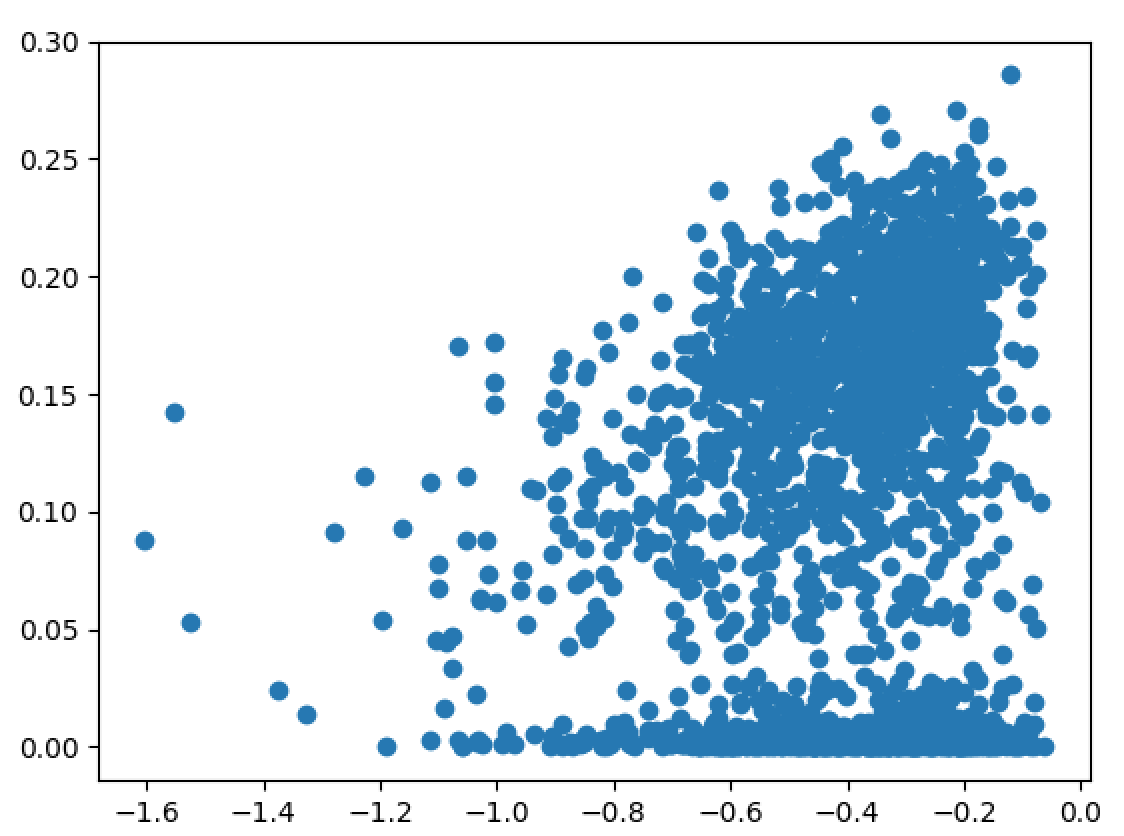}
        \caption{Blip2}
        \label{fig:sg}
    \end{subfigure} %
    
    \caption{The Y-axis shows the Hellinger's distance between the maximal and minimal marginals and the X-axis shows the negative log-sigmoid of the average similarity between CLIP embeddings for test images and human-written captions (closer to 0 is a better score).}
    \label{fig:bias}
\end{figure*}
}

\begin{figure*}
\centering
    \subfigure[SGAE]{\includegraphics[scale=0.25]{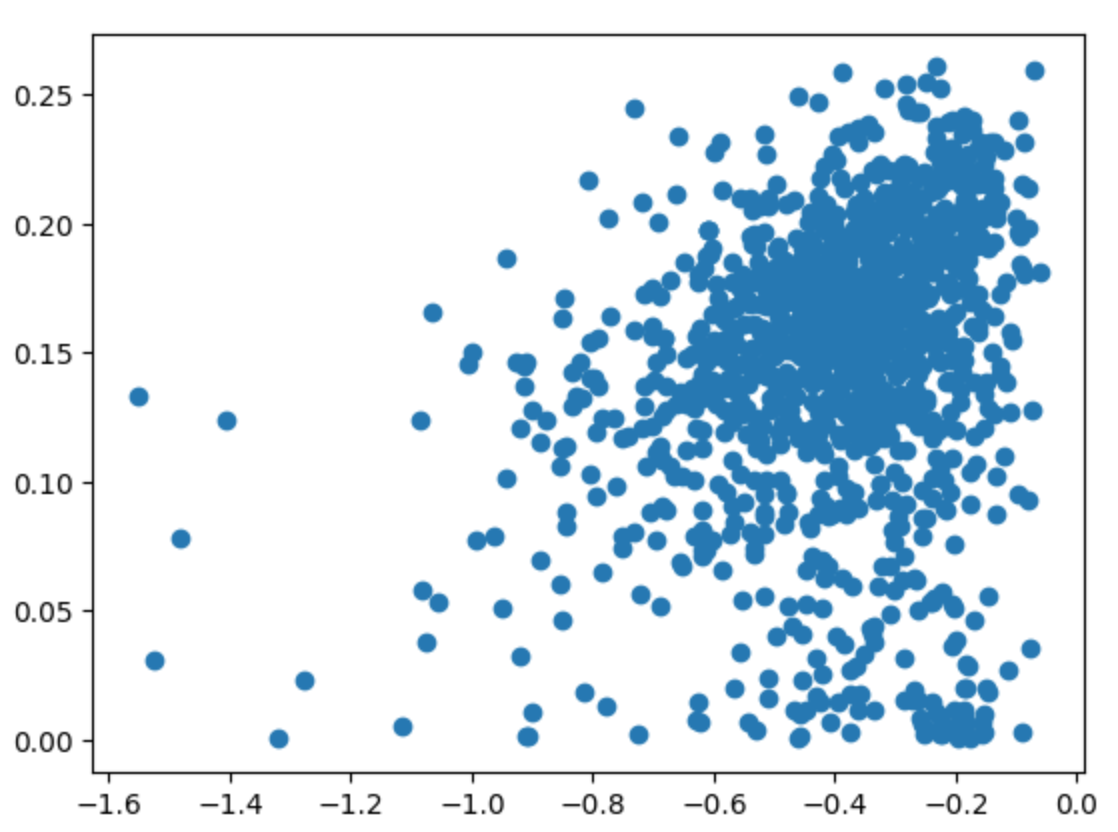}}
    \subfigure[AoA]{\includegraphics[scale=0.25]{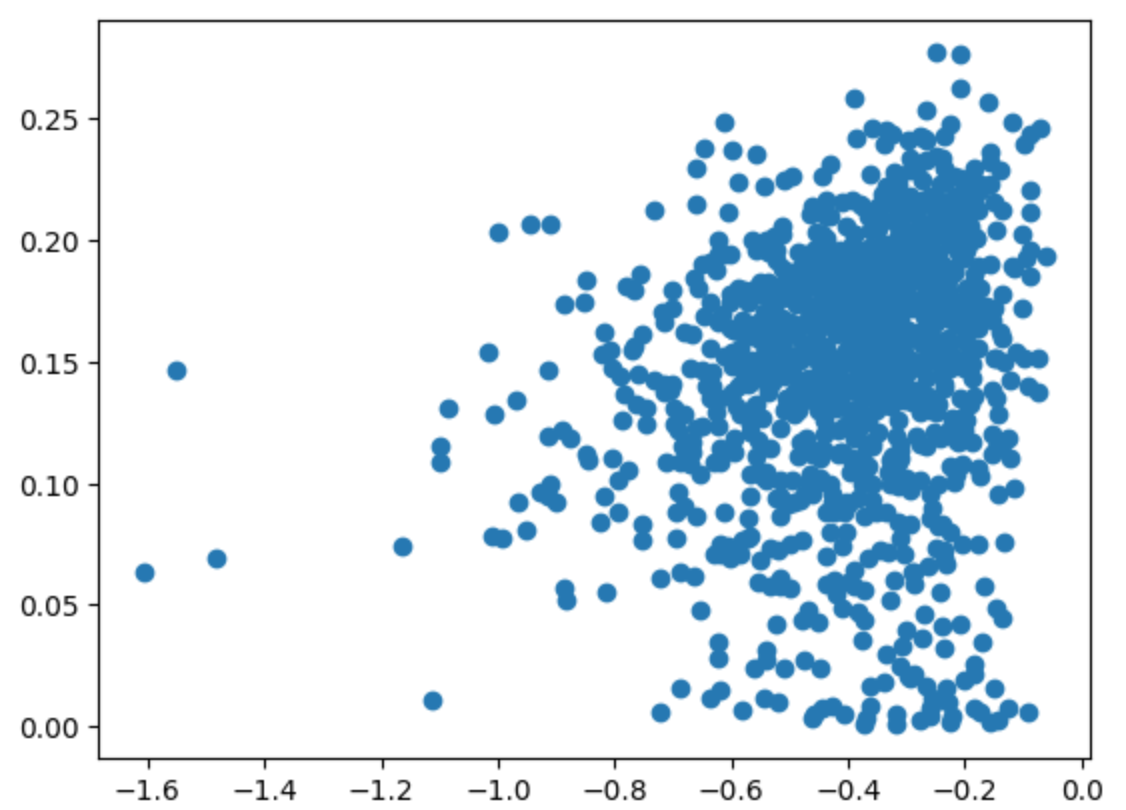}}
    \subfigure[M2]{\includegraphics[scale=0.25]{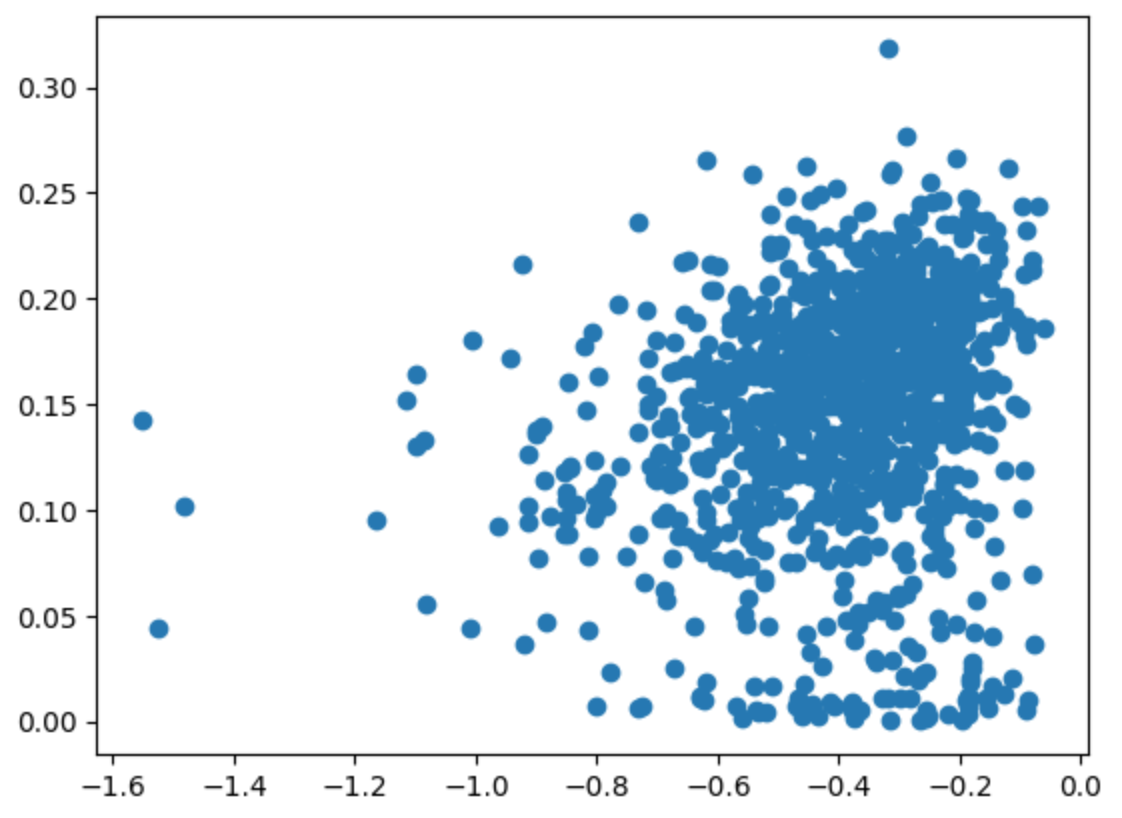}}
    \subfigure[XLAN]{\includegraphics[scale=0.25]{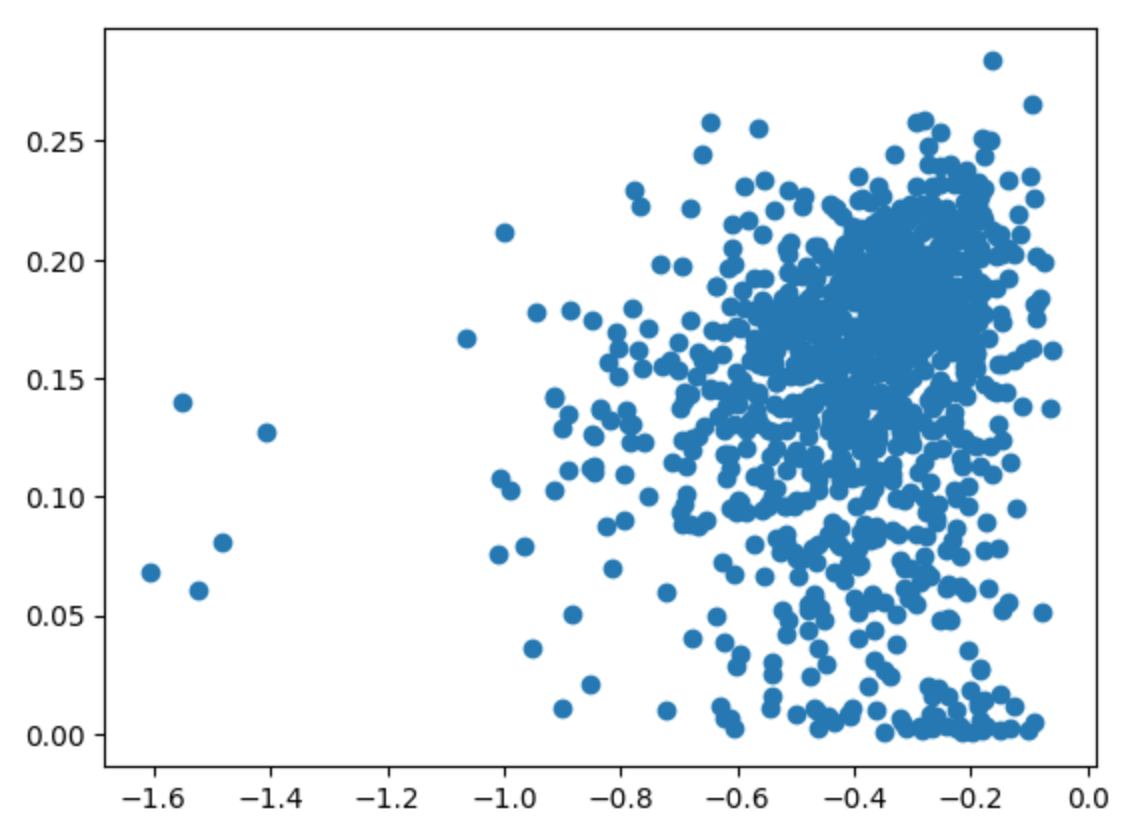}}
    \subfigure[BLIP2]{\includegraphics[scale=0.25]{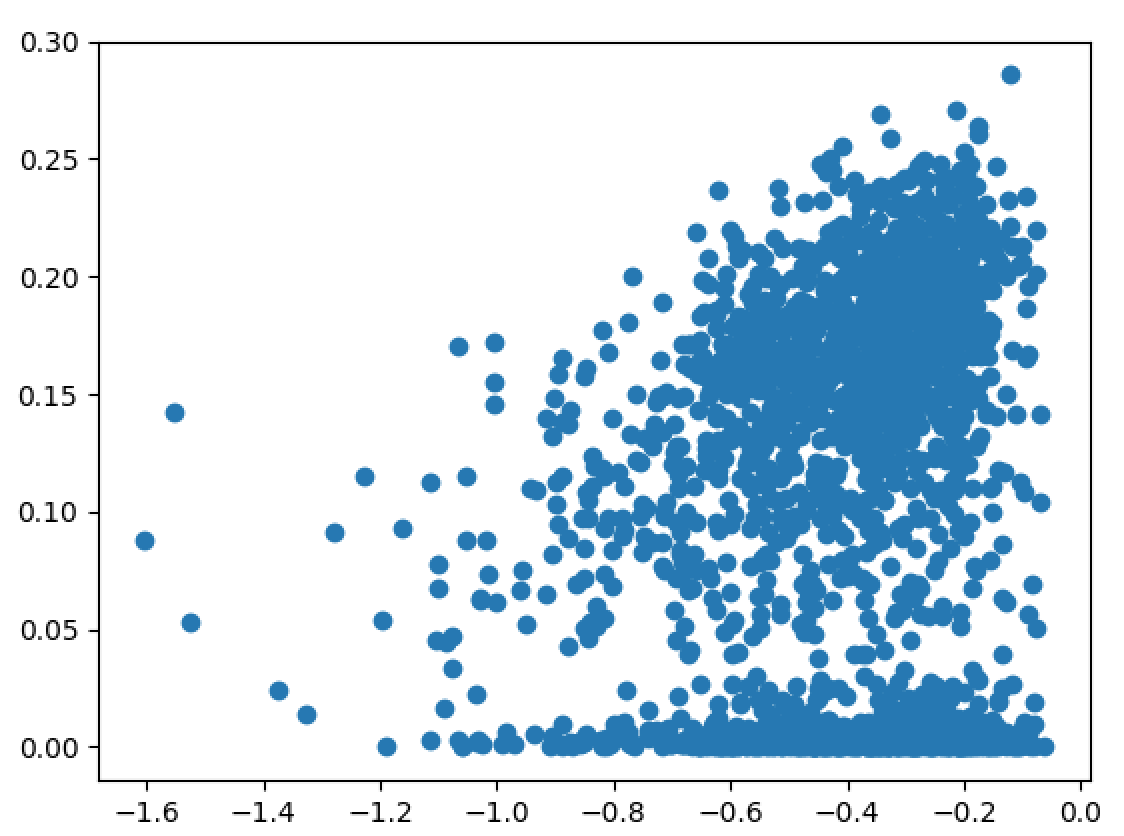}}
    \caption{\label{fig:bias}The Y-axis shows the Hellinger's distance between the distributions for contrastive examples and the X-axis shows the negative log-sigmoid of the average similarity between CLIP embeddings for test images and human-written captions (closer to 0 means semantically closer).}
\end{figure*}

\subsection{AMT Study}

%To evaluate the models based on the marginal density estimates using importance weighting, 

We evaluate if for a generated caption by a model we can sample training examples that can help support the knowledge required for its generation. To do this, we designed an AMT study as follows. For a given test image and caption generated by a model, we provided users with training examples with contrastive probabilities selected by our approach.
We asked users to rate on a Likert scale (1-5) if the explanation matched with their own perception of how the model generates its caption by learning concepts from the positive training example and contrasting it with similar concepts shown in the negative training example.

%with highest and lowest marginal values as examples of contrastive instances that the model most likely learns (or is least likely to learn) from to generate the caption for the test image. We asked users to rate on a Likert scale (1-5) if the learning bias of the model matched with their own perception. Some examples are illustrated in Fig.~\ref{fig:exampleamt}.
We used a total of 1000 AMT workers to obtain responses. Each explanation was given by 5 different workers. The results on captions generated from the 5 different models are shown in Fig.~\ref{fig:turkresults1}. 
For XLAN, the users were more sure (rating 5) that the presented examples supported the generated caption compared to BLIP2. This observation is consistent with the MAP results and shows that the use of LLMs makes it harder in BLIP2 to back-trace where it acquired its knowledge from.
%Overall, most users felt that the learning bias of the models matched with their own perception of how the model learns as indicated by the large number of 4/5 ratings across all models. 

%Further, the non-VLLM models showed better explainability than VLLMs which indicates that we were able to select examples from the training distribution to more accurately reflect a model's learning in the case of non-VLLM models. This illustrates that VLLMs use knowledge to generate their captions that cannot be easily explained to human users. The model that generated captions best explained by the examples (rating equal to 5) was XLAN which is also consistent with the results that we obtained using MAP inference.
We also performed the paired t-test to check if the difference in responses were statistically significant. Table~\ref{tab:ttest} shows these results and as we see, the response differences were statistically significant for most pairs except when we paired M2 captions with SGAE. Further, the responses for the BLIP2 model was significantly different from all the non-VLLM models which indicates the distinction between how VLLMs learn from the training examples compared to non-VLLM models. 

\begin{table}[!t]
    \centering
    \parbox{0.4\textwidth}{}{
    \caption{\label{tab:ttest} Paired t-test results, values in red were not statistically significant ($p>0.05$).}

    \begin{tabular}{|c|c|c|c|c|c|}
            \hline
		& \textbf{AoA} & \textbf{M2} & \textbf{XLAN} & \textbf{SGAE} & \textbf{Blip2}\\
            \hline
            \hline
            \textbf{AoA} & & 2.087 & 0.912 & 3.008 & 0.464\\
            \textbf{M2} & & & {0.78} & \textcolor{red}{-1.9} & 2.650\\
            \textbf{XLAN} & & & & 1.2 & 3.579\\
            \textbf{SGAE} & & & & & 1.286\\
		\hline
    \end{tabular}
    }
\end{table}

\eat{
\begin{table}[!t]
    \centering
    \parbox{0.35\textwidth}{!}{
 \caption{\label{tab:kappa1}Cohen kappa inter-rater agreement scores.}
    \begin{tabular}{|c|c|c|c|c|c|}
        \hline
        \textbf{..} & \textbf{$R_1$} & \textbf{$R_2$} & \textbf{$R_3$} & \textbf{$R_4$} & \textbf{$R_5$}\\
        \hline
        \hline
        \textbf{$R_1$} & 1 & 0.355 & 0.225 & 0.327 & 0.323 \\ 
        \textbf{$R_2$} &  & 1 & 0.334 & 0.355 & 0.285 \\
        \textbf{$R_3$} &  &  & 1 & 0.323 & 0.204  \\        
        \textbf{$R_4$} &  &  &  & 1 & 0.192 \\ 
        \textbf{$R_5$} &  &  &  &  & 1 \\
        \hline
    \end{tabular}
    }
\end{table}
}

\subsection{Analyzing the Distribution}

%We compare the difference between maximal and minimal marginal distributions that we infer on the training data with the similarity between image embeddings and text embeddings computed using CLIP on the reference captions. 
We analyze the difference between marginal probabilities of the contrasting training examples selected by our explainer.
Specifically, we compare this difference to the similarity between the test image embedding and the {\em ground-truth} (human-written) caption's text embeddings computed using CLIP.
%increased difference in distributions should correspond to increased semantic similarity between the embeddings. 
%This implies that when the image can be well-explained by captions (larger similarity between embeddings), the in
%To do this, we compute the average similarity between the image embedding and the embeddings for the 5 reference captions.
%We used  CLIPScore~\cite{hessel2021clipscore} on the ground-truth captions for the test data, i.e., captions written by humans and computed the average score over the 5 captions. 
To measure the distance between distributions, we use the Hellinger's distance~\cite{gonzalez2013class} (a symmetric distance between probability distributions) between the marginal probabilities for the contrastive examples. We expect that when the text and image have greater similarity, this implies that the training examples should be able to help learn concepts more reliably and therefore, the distance between distributions should be larger.
The results are shown in Fig.~\ref{fig:bias}. As seen in these results, out of all the models, for the VLLM BLIP2, the difference in distributions seems less explainable where in many test images, larger semantic similarity between the embeddings does not imply that the difference in distributions is large. Thus, while external sources help in caption generation with novel concepts, it does makes disentangling what is learned in fine-tuning harder.
%One reason for this could be that compared to other models, in VLLMs, a large contributor in its learning bias may be external datasets/knowledge.

%for all the models (to different extents) larger average similarity which is indicative that the image was semantically similar to the text also shows a larger distance between the distributions.

%Thus, the models tend to use this difference to 

%also implies that our approach in such cases will provide an explanation with examples from more diverse contexts (shown by the larger distance between the two distributions).

\section{Conclusion}

To better understand the knowledge acquired in fine-tuning versus that acquired during pre-training, we developed a probabilistic model (HMLN) that combines visual and symbolic features from the training instances used for fine-tuning. We then developed two inference tasks where we estimate if the generated caption can be supported by the training data distribution. We performed experiments on MSCOCO on different types of visual captioning models and showed that for BLIP2 which uses a LLM within the model, it can be harder to find support for its caption generation within the fine-tuning examples compared to other approaches that do not use LLMs. In future, we plan to generalize our work across different VLLMs and also across different multimodal tasks such as Visual Question Answering.

%We performed experiments on the MSCOCO dataset and showed that for models that have complex pretraining pipelines, the caption generation is not easily explainable from training examples. In future, we will apply our framework to explain other multimodal tasks such as VQA.
\bibliography{ref}

\end{document}